\PassOptionsToPackage{table}{xcolor}
\documentclass{article}
\usepackage{tactohand_preprint,times}
\usepackage{amsmath,amsfonts,bm}

\def\eqref#1{equation~\ref{#1}}
\def\1{\bm{1}}

\DeclareMathAlphabet{\mathsfit}{\encodingdefault}{\sfdefault}{m}{sl}
\SetMathAlphabet{\mathsfit}{bold}{\encodingdefault}{\sfdefault}{bx}{n}

\usepackage{amsmath,amssymb,booktabs,graphicx}
\usepackage{wrapfig}
\usepackage{tikz}
\usepackage{float}
\usepackage{placeins}
\usepackage{needspace}
\usepackage{multirow}
\usepackage{longtable}
\usepackage{siunitx}
\usepackage{xcolor}
\definecolor{tableblue}{HTML}{244A73}
\definecolor{tablefocus}{HTML}{FFF1D6}
\usepackage[hidelinks]{hyperref}
\usepackage{url}

\title{TactoHand: Learning Geometry-Based Dense Hand-Surface Interaction Fields from Video}
\author{\shortstack[c]{%
Ziheng Xi\textsuperscript{1}, Weibo Li\textsuperscript{2}, Jiayi Yu\textsuperscript{1}, Sipeng Zheng\textsuperscript{3}, Wanpeng Zhang\textsuperscript{3},\\[3pt]
Haiweng Xu\textsuperscript{2}, Hao Luo\textsuperscript{2}, Chuan Mao\textsuperscript{2}, Zongqing Lu\textsuperscript{2}\\[8pt]
\normalfont\textsuperscript{1}Tsinghua University\\
\normalfont\textsuperscript{2}Peking University\\
\normalfont\textsuperscript{3}BeingBeyond
}}

\iclrfinalcopy
\begin{document}
\maketitle
\begin{abstract}
Understanding how humans interact with the world requires a representation of physical interactions at the hand surface. 
Tactile sensing captures these interactions, but collecting tactile data at scale requires specialized hardware.
We introduce \textbf{TactoHand}, a model that predicts three dense hand-surface interaction fields---contact, proximity, and 3D force---from monocular video.
We derive geometry-based contact and proximity labels and motion-derived force labels from eligible training data within a curated collection spanning approximately 50 million source frames.
TactoHand uses a temporal transformer to aggregate visual evidence across frames, while three learned queries decode the corresponding hand-surface fields.
A differentiable physics layer further constrains force predictions using observed object motion.
Across in-domain and out-of-domain benchmarks, TactoHand improves contact and proximity estimation over prior methods.
In addition, physics supervision reduces position and rotation errors in motion rollouts, and the predicted contact and proximity fields improve downstream hand–object reconstruction.
Together, these results show a scalable, low-cost path to learning hand-surface interaction fields without collecting tactile measurements.
Code will be released at \url{https://github.com/zhenqis123/Tactohand}.

%demonstrate the value of modeling physical interactions directly on the hand surface.
%\textcolor{red}{For the first time, this allows us to at a low cost.}

%\textcolor{red}{harmonize} 24 hand–object, hand–hand, and hand–scene data sources totaling 50M frames, deriving high-fidelity, geometry-based contact and proximity labels across the collection and motion-derived force labels for the hand–object subset. 

\end{abstract}

\section{Introduction}
\setlength{\textfloatsep}{10pt}
Hands mediate human interactions with objects, the surrounding environment, and other hands.
Although the interacting entity and its geometry vary, the hand surface provides a common reference for describing where and how these interactions occur.
Contact, proximity, and force provide complementary geometric and physical cues that could support interaction reconstruction, action understanding, and embodied learning.
We call these vertex-wise quantities hand-surface interaction fields: contact describes where interaction occurs, proximity captures separation from the counterpart surface, and 3D force characterizes the hand's mechanical effect.
%Representing interaction on this surface could provide reusable physical cues for interaction reconstruction, action understanding, and embodied learning.
%We describe these cues as dense, aligned fields of contact, proximity, and 3D force on the hand surface.

Prior work has modeled subsets of these hand-surface quantities.
HACO predicts dense contact across diverse interaction types~\citep{jung2025dense}; InterField estimates hand--object distance~\citep{fan2023arctic}, HOPE jointly predicts contact and normal pressure from video~\citep{jeon2026hope}, and EgoPHI estimates contact and normal forces from egocentric images given object geometry~\citep{ilic2026egophi}.
These works motivate a shared hand-surface representation that brings contact, proximity, and 3D vertex forces together.
We therefore represent all three as dense, aligned fields on the canonical MANO hand surface.

Learning these fields from monocular video, however, requires supervision that covers diverse interaction types and physical quantities.
We harmonize 24 datasets comprising around 50 million frames, spanning hand--object, hand--hand, and hand--scene interactions.
We derive contact and proximity labels from the available geometry.
For eligible hand--object sequences, we additionally derive vertex-force labels from observed object trajectories.
Object motion constrains the aggregate force and torque, but does not uniquely determine how forces are distributed over the hand surface. We therefore use an SDF-aware formulation with geometric, frictional, and temporal constraints to obtain regularized vertex-force estimates.
%and geometry-based contact and distance proximity labels across the collection.
%For hand--object sequences with observed object motion, we also derive regularized vertex-force labels from trajectories and allocate them using an SDF-aware formulation.

\begin{figure*}[t]
    \centering
    \includegraphics[width=0.96\textwidth]{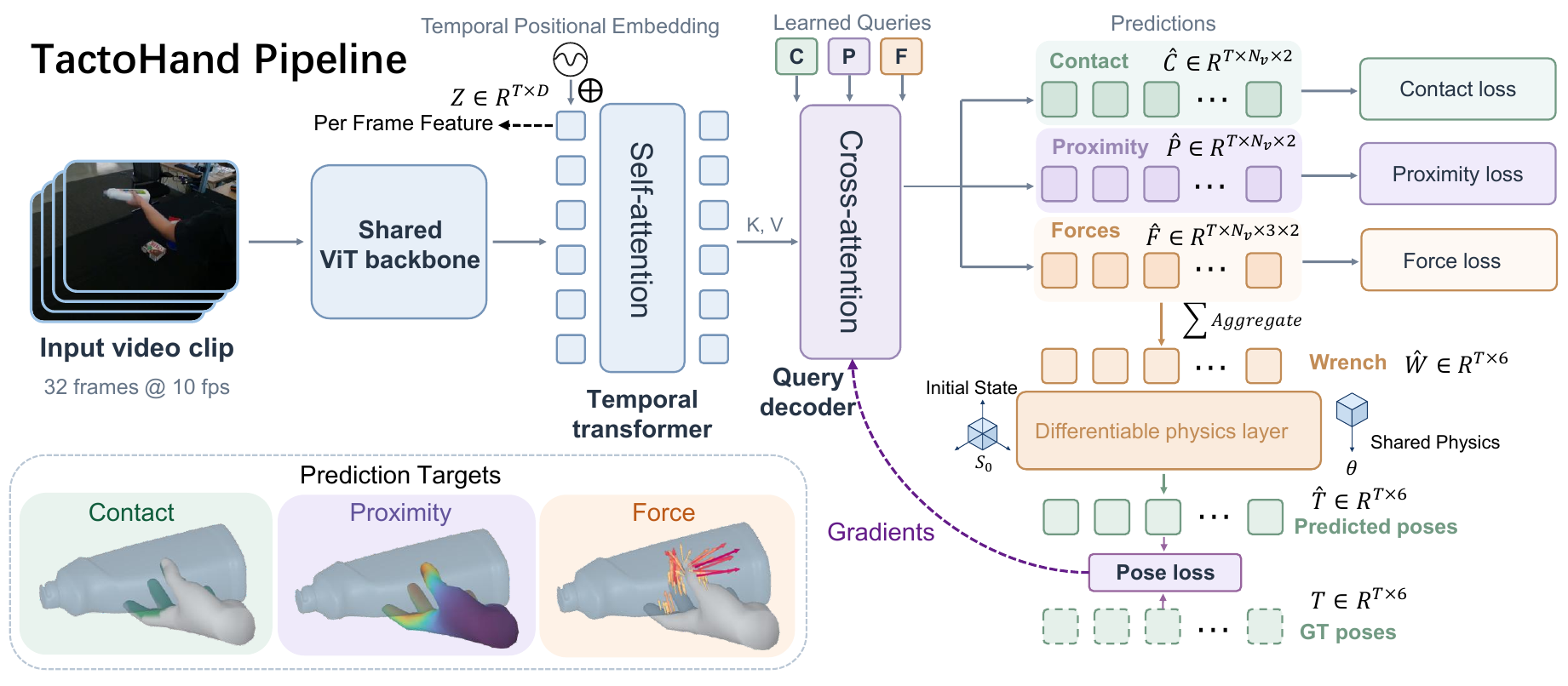}
    \caption{
    Overview of TactoHand. A shared video encoder extracts frame-level features, aggregated by a temporal transformer.
    Three learned queries decode the contact, proximity, and 3D force interaction fields at corresponding MANO vertices.
    Predicted vertex forces are aggregated into an object-level wrench for differentiable rigid-body supervision.
    Contact, proximity, force, and pose losses are applied according to the supervision available for each sample.
    }
    \label{fig:pipeline}
\end{figure*}

Inferring these fields remains challenging because contact regions are often occluded and individual frames provide limited evidence about interaction dynamics.
We introduce \textbf{TactoHand}, which uses a temporal transformer to aggregate visual evidence across frames and three learned queries to predict contact, proximity, and 3D force fields on corresponding MANO vertices, as shown in Figure~\ref{fig:pipeline}.
To connect vertex-level force predictions with object motion, we transform the predicted vertex forces into a common coordinate frame and analytically aggregate them into object-level wrenches.
For eligible rigid sequences, a differentiable dynamics layer uses these wrenches to propagate object motion and constrains force estimates through observed trajectories.
This motion-level supervision complements the vertex-level targets by encouraging predicted forces to explain the resulting object motion.
%Occlusion and ambiguous dynamics make these fields difficult to infer from monocular video.
%TactoHand aggregates temporal visual evidence to predict contact, proximity, and 3D force vectors on corresponding MANO vertices (Figure~\ref{fig:pipeline}).

We evaluate TactoHand on both interaction field prediction and downstream hand--object reconstruction.
TactoHand improves contact and proximity estimation across multiple benchmarks.
On HOCap, which is excluded from training, TactoHand achieves a contact Macro-F1 of 50.5, compared with 18.0 for HACO.
Physics supervision further reduces position and rotation errors in reference-assisted rigid-body rollouts.
In downstream hand--object reconstruction, the predicted contact and proximity fields provide geometric constraints that improve hand--object alignment and reconstruction accuracy.

Our contributions are:
\begin{itemize}
    \setlength{\itemsep}{2pt}
    \setlength{\parsep}{0pt}
    \setlength{\parskip}{0pt}
    \item A unified representation of hand-surface interaction that jointly models dense contact, proximity, and 3D vertex forces at corresponding vertices of the MANO mesh.
    \item TactoHand, a model that estimates these fields from monocular video, with differentiable rigid-body supervision to constrain vertex forces.
    \item A heterogeneous collection of 24 hand interaction datasets, with geometry-derived contact and proximity labels and motion-derived force estimates where applicable, supporting direct field evaluation and downstream hand--object reconstruction.
\end{itemize}

%remain largely studied in isolation.
%Contact indicates where physical interaction occurs, proximity describes surrounding spatial relationships, and forces characterize the mechanical effects produced by the hand.
%Although the interacting entity varies, ranging from objects and the surrounding environment to another hand, the hand surface provides a common coordinate system for describing interaction.

\setlength{\textfloatsep}{20pt plus 2pt minus 4pt}
\section{Related Work}
\label{sec:related-work}

\paragraph{Hand-surface Contact and Proximity. }
Grasp datasets and image-based reconstruction pipelines represent hand--object contact~\citep{brahmbhatt2020contactpose,tse2022s,hu2024explicit}. HACO predicts dense MANO contact across interaction types, and ContactPrompt predicts contact without training~\citep{jung2025dense,jung2026training}. InterField instead predicts continuous hand--object distance, while contact potential fields encode vertex relationships for pose refinement~\citep{fan2023arctic,yang2021cpf}. Related full-body work models human--scene contact or implicit human--object geometry~\citep{hassan2021posa,huang2022bstro,tripathi2023deco,xie2022chore}. TactoHand aligns contact and proximity on MANO vertices across hand--object, hand--hand, and hand--scene data.

\paragraph{Visual Prediction of Pressure and Force. }
Visual motion has been used to estimate interaction forces~\citep{li2021estimating}. PressureVision and EgoPressure learn from instrumented planar surfaces, whereas EgoTactile and TouchSight use glove measurements~\citep{grady2022pressurevision,zhao2024egopressure,zeng2026egotactile,zhou2026touchsight}. HOPE and EgoTac combine measured pressure or force with geometric contact on MANO vertices~\citep{jeon2026hope,zhang2026egotac}. EgoPHI predicts contact and normal-direction force given an egocentric image and object geometry~\citep{ilic2026egophi}. TactoHand derives three-component vertex-force targets from object motion for sequences with suitable object trajectories.

\paragraph{Motion and Physical Consistency.}
Prior work matches simulated contact forces to observed motion or couples them with hand--object reconstruction~\citep{ehsani2020force,hu2022physical}; ViTaM-D adds distributed tactile sensing~\citep{yu2024dynamic}. Differentiable rigid-body models allow motion errors to supervise predicted forces~\citep{qiao2020scalable,werling2021fast}. Since motion constrains only the aggregate wrench, TactoHand regularizes its vertex-force targets and applies a reference-assisted motion loss to eligible rigid sequences.

\paragraph{Interaction Cues for Downstream Tasks.}
Contact and distance constraints aid hand--object reconstruction~\citep{hasson2019learning,grady2021contactopt,hasson2021towards,qi2024hoisdf}. Hand affordances and measured glove forces have also supported robot learning and physical action understanding~\citep{srirama2024hrp,dessalene2026feel}. We test downstream utility by adding predicted contact and proximity as fixed constraints to existing reconstruction pipelines.
%These findings motivate learning interaction cues that can support subsequent perception and learning tasks.

%by  contact and proximity fields improve existing hand--object reconstruction methods.

\section{Problem Formulation}
\label{sec:formulation}

Let $I_{1:T}$ denote a monocular RGB clip of $T$ frames sampled at 10\,Hz.
Each hand is represented by the fixed MANO topology with $V=778$ vertices \citep{romero2017embodied}. 
We index each vertex by $i=(h,v)$, where $h\in\{L,R\}$ denotes the hand side and $v\in\{1,\ldots,V\}$ denotes the vertex index.
TactoHand estimates three aligned hand-surface interaction fields on these vertices:
\begin{equation}
\Phi_\theta(I_{1:T})=\bigl\{(\widehat c_{t,i},\widehat r_{t,i},\widehat{\boldsymbol\ell}_{t,i})\bigr\}_{t,i},
\quad \widehat c_{t,i},\widehat r_{t,i}\in[0,1],\quad \widehat{\boldsymbol\ell}_{t,i}\in\mathbb R^3.
\label{eq:vertex-field-prediction}
\end{equation}
Here, $\widehat c_{t,i}$ denotes the contact probability,
%where interaction occurs; 
while $\widehat r_{t,i}$ measures proximity to the counterpart surface, with larger values indicating greater closeness.
$\widehat{\boldsymbol\ell}_{t,i}$ contains raw vertex-local force components predicted from RGB. Converting the gated components into camera-frame forces $\widehat{\mathbf f}_{t,i}$ additionally requires ground-truth posed MANO geometry (Equation~\ref{eq:camera-vertex-force}).
Force supervision is applied only to eligible hand–object samples. 
For hand–hand and hand–scene samples, force losses are masked out, while contact and proximity supervision is retained wherever valid annotations are available.
%Hand--hand and hand--scene samples mask force supervision but retain valid contact and proximity supervision.

\section{Dataset Construction}
\label{sec:label-generation}

\subsection{Data Sources}
\label{sec:collection}

We combine $24$ sources spanning diverse hand interaction scenarios, comprising $282{,}874$ curated sequences and $49.02$ million captured frames across camera views, approximately $5.2\times$ HACO’s frame count. 
For all datasets, we derive geometry-based contact and proximity labels on a shared set of MANO vertices. 
We additionally derive motion-based vertex-force labels for 15 hand--object datasets, including $13$ with rigid objects and two with articulated objects. These are collection counts; training excludes GigaHands force targets and all HOCap data.
Appendix Table~\ref{tab:dataset-inventory} reports the sequence and frame counts, together with the available supervision, for each source.

\subsection{Contact and Proximity Labeling}

For a hand vertex $\mathbf{p}$, we compute its unsigned distance $u$ to the original object surface in the object’s coordinate frame.
We then obtain the signed distance $s$ using a multiray inside–outside test, described in Appendix~\ref{app:inside-outside-test}, retaining the distance magnitude and assigning a negative sign to vertices inside the object and a nonnegative sign otherwise.
% \begin{equation}
% s=\begin{cases}
% -u, & \mathbf{p}\text{ is inside the object},\\
% \phantom{-}u, & \text{otherwise}.
% \end{cases}
% \label{eq:signed-distance-label}
% \end{equation}
For hand--hand and hand--scene data, we query the other posed hand or the available scene surface, respectively. When the counterpart surface is open and a reliable inside–outside classification is unavailable, we use the unsigned nearest-surface distance and disable inside-mesh inclusion.
The proximity target is defined as:
\begin{equation}
r=\exp\!\left(-\tau\min\{\max(s,0),d_{\max}\}\right).
\label{eq:proximity-label}
\end{equation}
This target equals one on and inside the surface, decreases with separation outside the surface, and remains constant beyond $d_{\max}$.
We use $\tau=20\,\mathrm{m}^{-1}$ and $d_{\max}=0.25$\,m. We assign contact labels using the following rule.

\begingroup
\setlength{\intextsep}{4pt}
\setlength{\columnsep}{6pt}
\setlength{\abovecaptionskip}{4pt}
\Needspace{16\baselineskip}
\begin{wrapfigure}[10]{r}{0.49\textwidth}
    \vspace{-8pt}
    \centering
    \includegraphics[width=\linewidth]{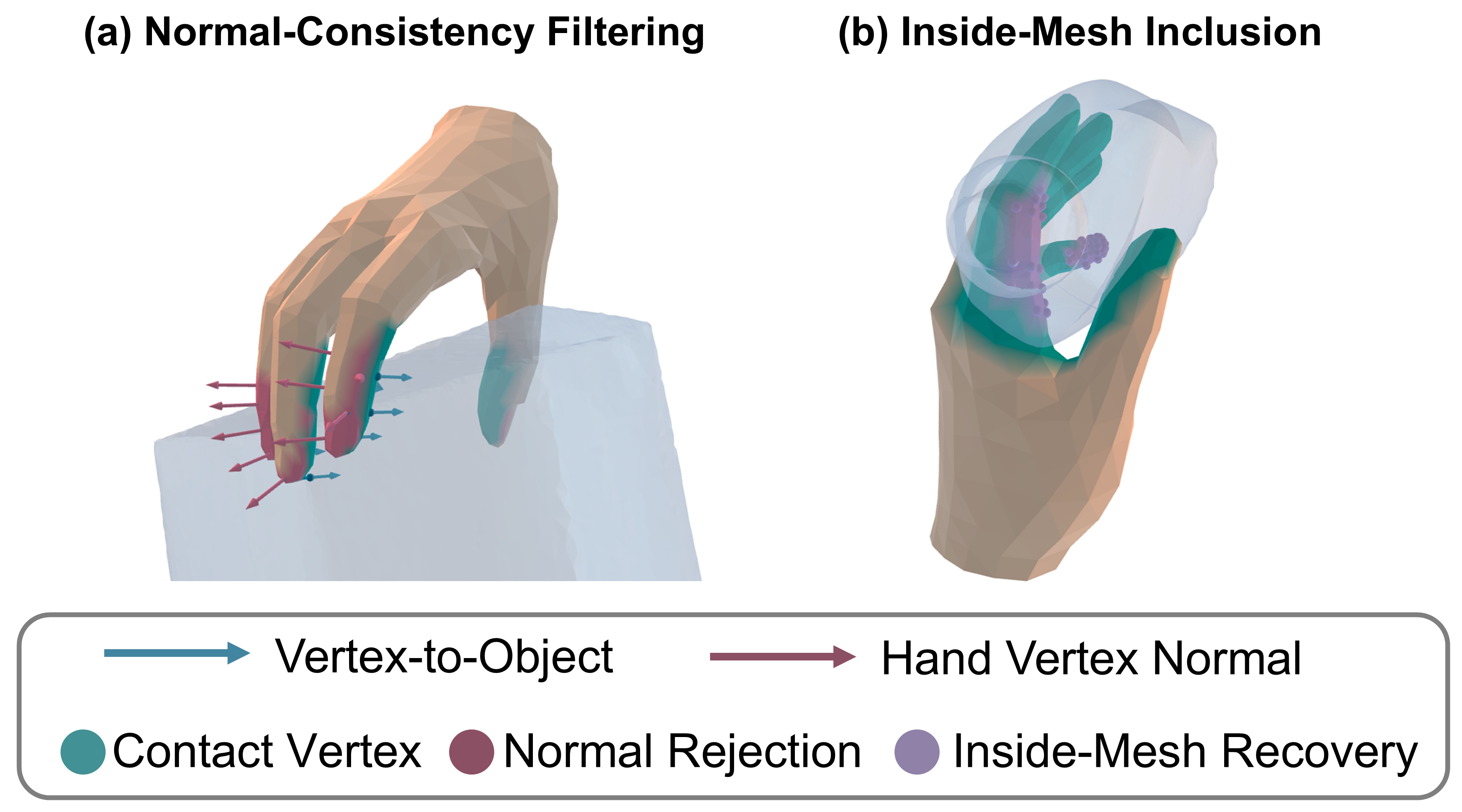}
    \caption{Contact-label construction: normal filtering (left) and inside-mesh inclusion (right).}
    \label{fig:contact-generation}
\end{wrapfigure}

\paragraph{Contact Decision.}
Let $\delta_D$ be the dataset-specific distance threshold, $\mathbf q$ the closest point on the counterpart surface, and $\mathbf n$ the outward hand-surface normal.
For an outside vertex with $u>0$, let $\kappa=\mathbf n^\top(\mathbf q-\mathbf p)/u$ measure normal alignment. For a closed object mesh, we assign contact by
\begin{equation}
c(\mathbf p)=\begin{cases}
1, & \mathbf p\text{ is inside the object},\\
1, & \text{outside},\ u\leq\delta_D,\ u<3\,\mathrm{mm},\\
1, & \text{outside},\ 3\,\mathrm{mm}\leq u\leq\delta_D,\ \kappa\geq0,\\
0, & \text{otherwise}.
\end{cases}
\label{eq:contact-decision}
\end{equation}
Here, ``outside'' includes on-surface vertices ($u=0$), which are labeled as contact without evaluating $\kappa$; the $3$\,mm band likewise bypasses the normal test where the closest-point direction is unstable.
Inside-mesh inclusion retains potential contacts displaced into the object by pose error.
For hand--hand and hand--scene data, we apply the same distance and normal tests, using inside-mesh inclusion only when a reliable sign is available.
Appendix~\ref{app:contact-label-comparison} provides the distance thresholds and a visual comparison of the labeling rules.
HACO assigns contacts by thresholding unsigned nearest-surface distance~\citep{jung2025dense}.
Our geometric corrections provide more faithful contact supervision, as illustrated in Appendix Figure~\ref{fig:Annotation-Comparison}.
\par
\endgroup

\subsection{Motion-Derived Force Labels}
\label{sec:force-label-generation}

The observed object motion determines the net wrench acting on the object, but does not uniquely determine the distribution of forces over the hand surface.
We therefore construct force supervision in two stages: we first recover an object-level target wrench from motion and then distribute it over plausible hand vertices under geometric, frictional, and temporal constraints.

For rigid objects, we smooth the observed pose trajectories at $30$\,Hz and estimate linear and angular accelerations. Newton--Euler inverse dynamics then gives the non-gravitational external wrench that must be explained by the hands and any support from the environment:
\begin{equation}
\mathbf{w}^{\star}_t =
\begin{bmatrix}
m(\mathbf{a}_t-\mathbf{g}) \\
\mathbf{R}_t\!\left(\mathbf{I}\boldsymbol\alpha_t+
\boldsymbol\omega_t\times\mathbf{I}\boldsymbol\omega_t\right)
\end{bmatrix},
\qquad
\mathbf{w}^{\mathrm{hand}}_t+\mathbf{w}^{\mathrm{env}}_t
=\mathbf{w}^{\star}_t.
\label{eq:inverse-dynamics-wrench}
\end{equation}
Here, $\mathbf R_t$ is the object orientation, $m=1\,\mathrm{kg}$ is the reference object mass and $\mathbf I$ is the body-frame inertia scaled to this mass.
The linear acceleration $\mathbf a_t$ and gravity $\mathbf g$ are expressed in the solver world frame with $\lVert\mathbf g\rVert=9.81\,\mathrm{m/s^2}$, while the angular velocity $\boldsymbol\omega_t$ and acceleration $\boldsymbol\alpha_t$ are expressed in the body frame.
We use dataset-provided gravity directions when available; otherwise, we estimate them from UprightNet RGB predictions~\citep{xian2019uprightnet} or scene geometry. 
%and $\mathbf g$ is gravity in the solver world frame, $\mathbf R_t$ is the object orientation, and . Here $\mathbf a_t$ is the linear acceleration, while $\boldsymbol\omega_t$ and $\boldsymbol\alpha_t$ are the body-frame angular velocity and acceleration. We use dataset-provided gravity directions when available; otherwise, we estimate them from UprightNet RGB predictions~\citep{xian2019uprightnet} or scene geometry. All directions are expressed in the solver world frame, with $\lVert\mathbf g\rVert=9.80665\,\mathrm{m/s^2}$.

\begin{figure}[H]
    \centering
    \includegraphics[width=0.75\textwidth]{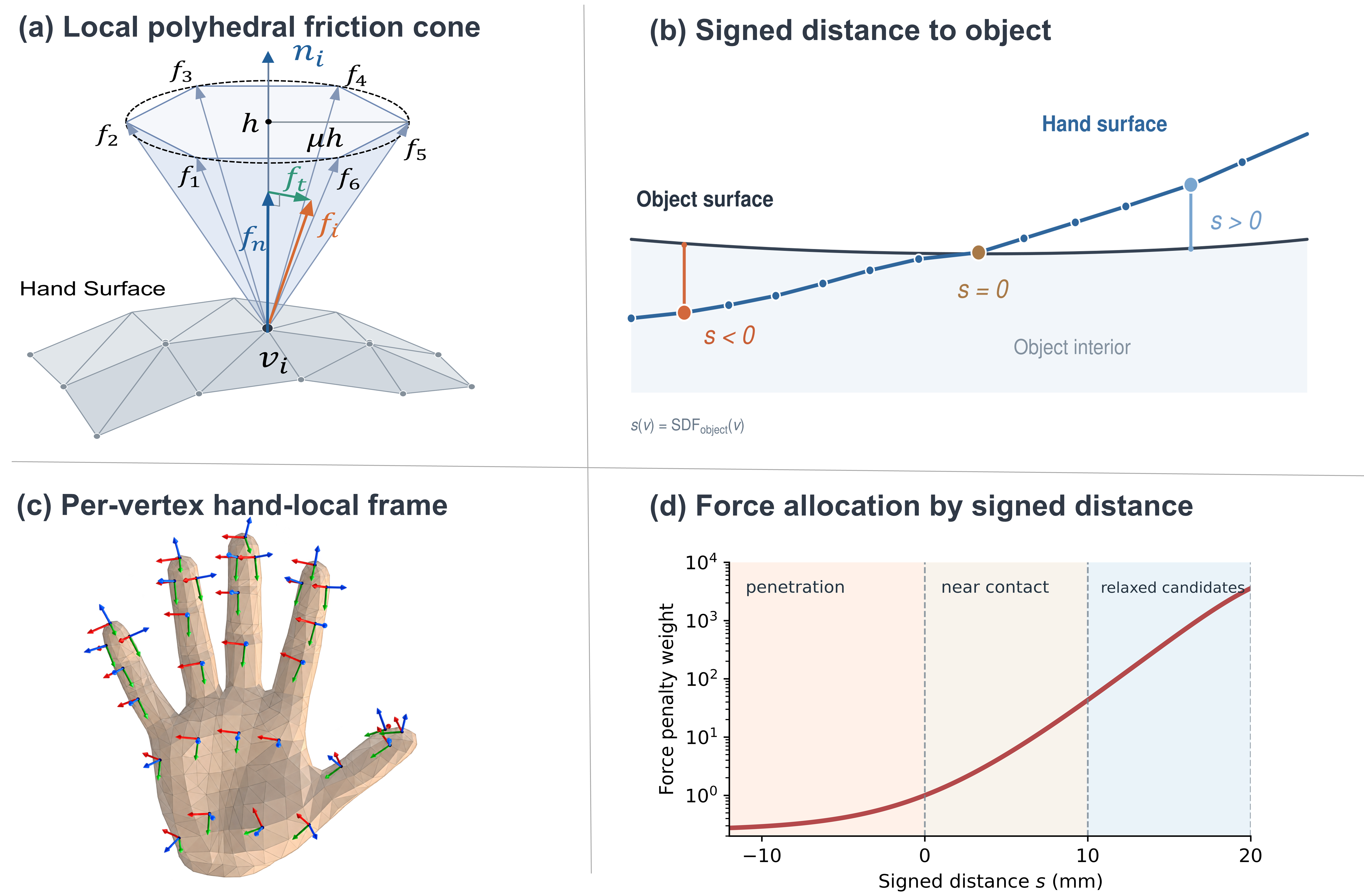}
    \caption{Force allocation geometry and weighting. Upper left: a local polyhedral friction cone constrains each active vertex force. Upper right: signed distance $s$ distinguishes penetrated ($s<0$), contacting ($s=0$), and separated ($s>0$) vertices. Lower left: vertex-local force directions on the hand surface. Lower right: the penalty weight $(q(s)^2+\epsilon)^{-1}$ rises toward the relaxed $20$ mm candidate cut-off.}
    \label{fig:force-allocation}
\end{figure}

The target wrench alone does not uniquely determine the force at each hand vertex. We resolve this ambiguity using the local hand--object geometry illustrated in Figure~\ref{fig:force-allocation}. Vertices with signed distance $s_{t,i}<20$\,mm are treated as force candidates, and each candidate force is constrained by a local friction cone defined by the hand-surface normal and tangent directions. We further bias the allocation toward vertices that are closer to, or penetrating, the object through
\begin{equation}
q_{t,i}=\frac{2}{1+\exp(s_{t,i}/\ell)},
\qquad \ell=4\ \mathrm{mm}.
\label{eq:force-candidate-priority}
\end{equation}
Because the force penalty is weighted by $(q_{t,i}^2+\epsilon)^{-1}$, vertices closer to or penetrating the object incur a lower cost.

We then solve for the hand forces jointly across the sequence. To account for interactions with supporting surfaces, we additionally introduce an upward environment force with magnitude $e_t\geq0$, wrench map $\mathbf G_t^{\mathrm{env}}$, and regularizer $\mathcal R_{\mathrm{env}}(e_{1:T})$. The resulting sequence-level convex program is
\begin{equation}
\begin{aligned}
\min_{\{\mathbf{f}_{t,i}\},\,\{e_t\}}
&\quad \lambda_f\sum_{t,i}\frac{\lVert\mathbf{f}_{t,i}\rVert_2^2}{q_{t,i}^2+\epsilon}
+\lambda_{\Delta f}\sum_{t,i}\lVert\mathbf{f}_{t+1,i}-\mathbf{f}_{t,i}\rVert_2^2
+\mathcal{R}_{\mathrm{env}}(e_{1:T})\\
\text{s.t.}&\quad
\sum_i\mathbf{G}_{t,i}\mathbf{f}_{t,i}+\mathbf{G}^{\mathrm{env}}_t e_t
=\mathbf{w}^{\star}_t,\\
&\quad \mathbf{f}_{t,i}\in\mathcal{K}(\mathbf{n}_{t,i},\mu),
\quad \mathbf{f}_{t,i}=\mathbf{0}\ \text{for inactive }i,
\quad e_t\geq0,
\end{aligned}
\label{eq:force-allocation}
\end{equation}
where $\mathbf G_{t,i}$ maps a vertex force to its force-and-torque contribution about the object center of mass, and $\mathcal K$ is a six-facet polyhedral approximation of the unilateral Coulomb friction cone. The equality constraint ensures that the allocated hand and environment forces reproduce the motion-derived target wrench, while the magnitude and temporal terms select a spatially and temporally regularized solution among the feasible allocations.

For articulated objects, we also account for joint forces and torques, mapping each hand contact to the object part it acts on (Appendix~\ref{app:articulated-force-labels}). Articulated sequences provide vertex-force supervision but are excluded from the rigid rollout loss. Appendix~\ref{app:force-label-details} provides further details on trajectory processing, rigid-force allocation and validation, and the articulated extension.

\raggedbottom
\section{TactoHand}
\label{sec:method}

\subsection{Video Encoding and Interaction-Field Decoding}

Each frame is encoded by a pretrained ViT-L DINOv3 backbone~\citep{simeoni2025dinov3}. A temporal transformer aggregates clip-level visual information, and three task-specific queries decode the hand-surface interaction fields for both hands at each 10\,Hz image timestamp. Architecture and input settings are provided in Appendix~\ref{app:network-architecture}.

\subsection{Dense Contact and Proximity}

Let $\widehat a_{t,h}$ denote the probability that hand $h$ makes any contact in frame $t$ (engagement), and $\widehat b_{t,h,v}$ the vertex contact probability conditioned on that event. We factor the final contact probability as
$\widehat c_{t,h,v}=\widehat a_{t,h}\widehat b_{t,h,v}$,
with engagement target
$a^\star_{t,h}=\mathbb{1}[\sum_v c^\star_{t,h,v}>0]$
on valid hand frames. Binary cross-entropy supervises engagement, while HACO's vertex-level class-balanced (VCB) loss~\citep{jung2025dense} supervises conditional contact on engaged frames. A separate proximity head is supervised with masked mean squared error against the targets defined in Equation~\ref{eq:proximity-label}.

The same engagement scalar $\widehat a_{t,h}$ also gates all vertex forces of hand $h$. This soft gate allows force and motion losses to propagate into both the engagement and force branches while encouraging zero force on no-contact hand frames.

\subsection{Vertex Forces and Analytic Wrench Aggregation}

Omitting the time index, let $\widehat\ell_{i,1},\widehat\ell_{i,2},\widehat\ell_{i,n}$ denote raw local force components along two tangents and the outward normal. Ground-truth posed MANO geometry defines this basis in camera coordinates. With the engagement gate for hand $h(i)$, the hand-on-object force is
\begin{equation}
\widehat{\mathbf f}_i=\widehat a_{h(i)}\left(\widehat\ell_{i,1}\mathbf t_{i,1}+\widehat\ell_{i,2}\mathbf t_{i,2}+\widehat\ell_{i,n}\mathbf n_i\right).
\label{eq:camera-vertex-force}
\end{equation}
The basis vectors and $\widehat{\mathbf f}_i$ are therefore expressed in the camera frame. Predicting force components relative to the hand surface reduces sensitivity to camera orientation (Figure~\ref{fig:force-allocation}, lower left; Appendix~\ref{app:field-clock-validity}).

At each image timestamp, the point-force loss $\mathcal{L}_{\mathrm{point}}$ supervises the reconstructed camera-frame forces while balancing active and near-zero targets (Appendix~\ref{app:field-clock-validity}). Given the object center of mass $\mathbf{x}$ and vertex positions $\mathbf{p}_i$ in the same frame, the corresponding hand wrench is obtained analytically as
\begin{equation}
\mathbf{w}^{\mathrm{hand}}
=
\begin{bmatrix}
\sum_i\widehat{\mathbf{f}}_i \\
\sum_i(\mathbf{p}_i-\mathbf{x})\times\widehat{\mathbf{f}}_i
\end{bmatrix}.
\label{eq:wrench-aggregation}
\end{equation}

\subsection{Motion-Consistent Rollout}

We further constrain the predicted forces through short-horizon rigid-body motion. Each 10\,Hz visual interval contains three 30\,Hz force intervals indexed by $s\in\{0,1,2\}$. At $s=0$, we use the image-aligned network prediction $\widehat{\mathbf f}_{t,i}$; at $s\in\{1,2\}$, we use the detached motion-derived reference $\operatorname{stopgrad}(\mathbf f^{\star}_{t,s,i})$. This keeps the image-aligned prediction differentiable while avoiding unconstrained force extrapolation between visual observations.

\begin{figure}[!t]
    \centering
    \includegraphics[width=0.85\textwidth]{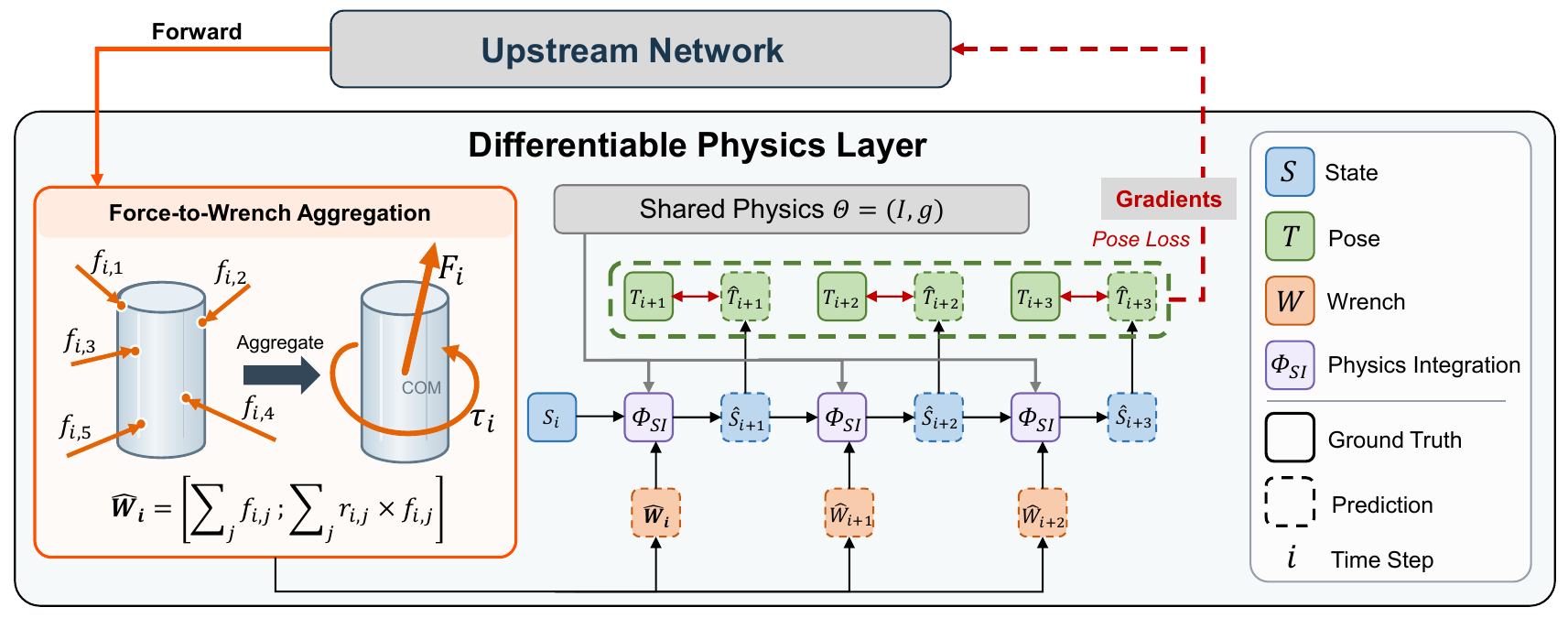}
    \caption{Differentiable rigid-body supervision. Indices $i,i+1,i+2$ are 10\,Hz image timestamps; each transition contains three 30\,Hz force intervals. Predicted vertex forces determine the object wrench at the image-aligned interval, while the two intermediate intervals use detached motion-derived reference forces. Pose errors supervise the propagated state.}
    \label{fig:differentiable-physics}
\end{figure}

Starting from the observed object state, we aggregate the vertex forces into a hand wrench and transform it from camera coordinates into the solver world frame. We then combine it with gravity and the estimated environment-support terms and integrate the resulting rigid-body dynamics. Because this integration is differentiable, motion-level errors propagate back through wrench aggregation to the predicted vertex forces.

Let $k$ index object states at the boundaries of successive 30\,Hz force intervals, with $k=0$ denoting the initial state. Let $\mathcal{H}$ denote the indices included in the short-horizon objective. We compare the propagated and observed object position and orientation:
\begin{equation}
\mathcal{L}_{\mathrm{roll}}
=
\frac{1}{|\mathcal H|}
\sum_{k\in\mathcal H}
\left(
\lambda_x
\lVert\widehat{\mathbf{x}}_k-\mathbf{x}^{\star}_k\rVert_2^2
+
\frac{\lambda_R}{2}
\lVert\widehat{\mathbf{R}}_k-\mathbf{R}^{\star}_k\rVert_F^2
\right).
\label{eq:rollout-pose-loss}
\end{equation}
The differentiable state update is detailed in Appendix~\ref{app:field-clock-validity} (Equation~\ref{eq:rollout-state-update}). The rollout loss is applied only to eligible rigid sequences for which the aggregated hand force can be attributed to the target object; articulated and otherwise ineligible sequences retain their available field-level supervision.

\subsection{Training Objective}

Each training sample activates only losses supported by its annotations. We combine the weighted engagement and conditional vertex losses into $\mathcal{L}_{\mathrm{contact}}$ and define the overall training objective as
\begin{equation}
\mathcal{L}
=
\lambda_c\mathcal{L}_{\mathrm{contact}}
+
\lambda_d\mathcal{L}_{\mathrm{prox}}
+
\lambda_p\mathcal{L}_{\mathrm{point}}
+
\lambda_r\mathcal{L}_{\mathrm{roll}}.
\label{eq:capability-objective}
\end{equation}
Geometry-only sequences use contact and proximity supervision. Sequences with valid force labels additionally use $\mathcal{L}_{\mathrm{point}}$, while eligible rigid sequences also activate $\mathcal{L}_{\mathrm{roll}}$. Missing force supervision is masked. Repeated frames used to pad clips are excluded from force and rollout losses (Appendix~\ref{app:field-clock-validity}).

\section{Experiments}
\label{sec:experiments}

\subsection{Dense Proximity and Contact Comparison}

\begin{figure}[!t]
    \centering
    \includegraphics[
        width=\linewidth,
        trim=11bp 18bp 55bp 28bp,
        clip
    ]{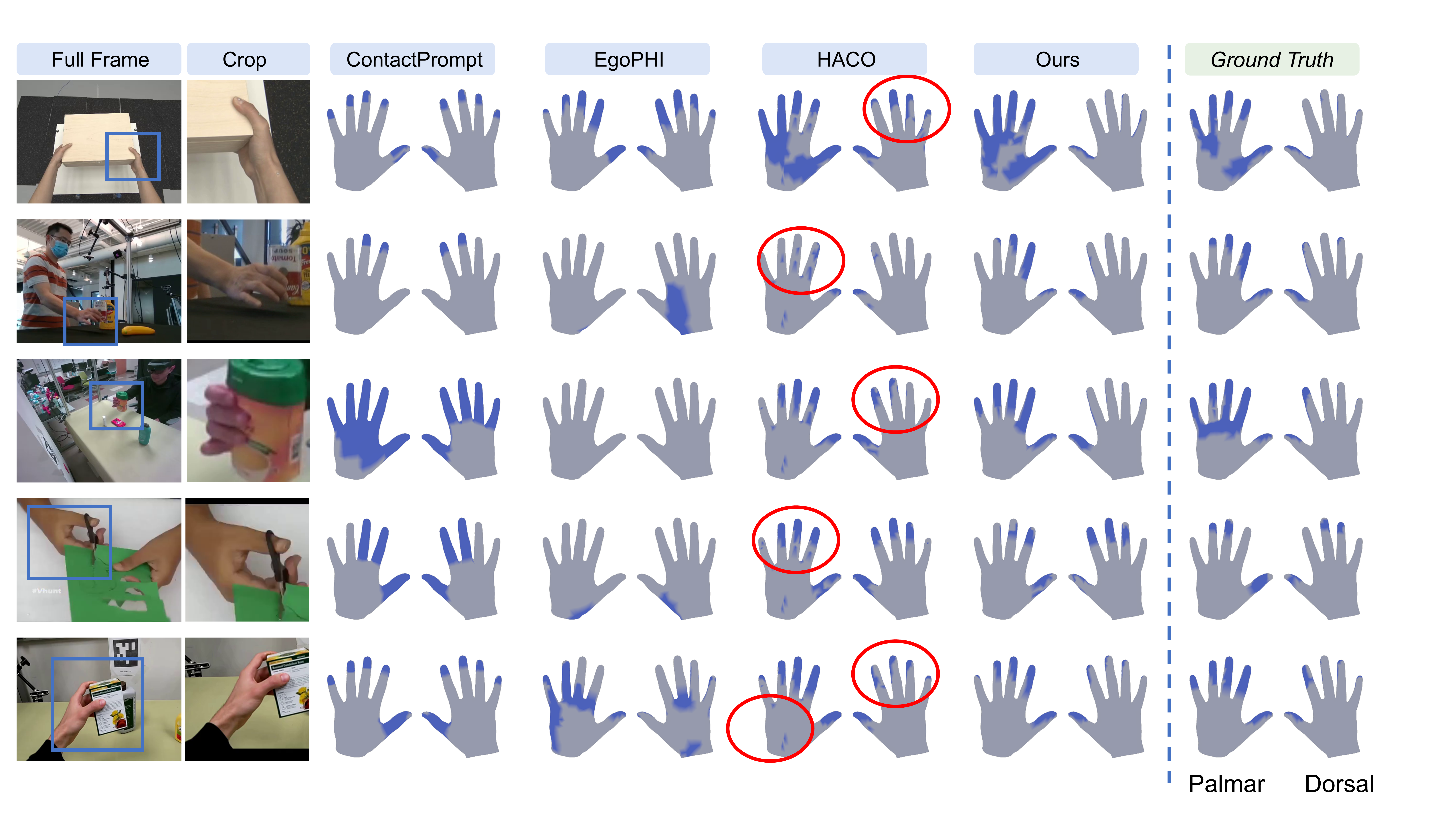}
    \caption{
    Contact comparison showing the input scene, hand crop, predicted contact maps, and geometry-derived reference.
    }
    \label{fig:qualitative-contact}
\end{figure}

Table~\ref{tab:prox_contact} compares TactoHand with prior proximity and contact predictors. We re-score released baseline checkpoints against the same annotations. For proximity, we compare against InterField-SF and InterField-LSTM~\citep{fan2023arctic}. For contact, we compare against HACO~\citep{jung2025dense}, EgoPHI~\citep{ilic2026egophi}, and ContactPrompt~\citep{jung2026training}. EgoPHI uses single RGB frames together with solver-derived MANO vertices and target-object masks and meshes. ContactPrompt is evaluated quantitatively only on MOW because full evaluation requires a separate VLM prompt for each frame; elsewhere, we include it qualitatively in Figure~\ref{fig:qualitative-contact}. Input settings, dataset splits, and the frames and vertices included in each metric are detailed in Appendix~\ref{app:evaluation-details}.

We evaluate cross-dataset transfer on HOCap, which is excluded from training.

\begin{table}[!t]
\centering
\footnotesize
\caption{
Proximity (left) and contact (right). ADE is in mm, Accel. in m/s$^2$, and P/R/F1 in percent. ADE-100 clips predictions and targets at 100\,mm; Macro-F1 averages F1 over frame--hand pairs with ground-truth or predicted contact. Bold marks the best per dataset. $\ddagger$ HOCap is excluded from training. See Appendix~\ref{app:evaluation-details} for metrics.
}
\label{tab:prox_contact}
\setlength{\tabcolsep}{2.5pt}
\renewcommand{\arraystretch}{1.10}
\resizebox{\linewidth}{!}{%
\begin{tabular}{@{}ll S[table-format=3.2] S[table-format=2.2] S[table-format=1.2] @{}!{\hspace{6pt}} l S[table-format=2.1] @{\hspace{4.5pt}} S[table-format=2.1] @{\hspace{4.5pt}} S[table-format=2.1] @{\hspace{4.5pt}} S[table-format=2.1]@{}}
\toprule
& \multicolumn{4}{c}{\textsc{Proximity}}
& \multicolumn{5}{c}{\textsc{Contact}} \\
\cmidrule(lr){2-5}
\cmidrule(l){6-10}
Dataset
& Method
& {ADE $\downarrow$}
& {ADE-100 $\downarrow$}
& {Accel. $\downarrow$}
& Method
& {P $\uparrow$}
& {R $\uparrow$}
& {Macro-F1 $\uparrow$}
& {Micro-F1 $\uparrow$} \\
\midrule

HOCap$^\ddagger$
& InterField-SF & 132.51 & 38.15 & 1.06
& HACO & 11.1 & \bfseries 77.7 & 18.0 & 19.5 \\

& InterField-LSTM & 134.69 & 40.10 & \bfseries 0.47
& EgoPHI & 10.3 & 37.3 & 11.4 & 16.2 \\

& \textcolor{tableblue}{Ours (SF)}
& 46.09 & \cellcolor{tablefocus}\bfseries 6.35 & 1.35
& \textcolor{tableblue}{Ours (SF)}
& \bfseries 68.9 & 57.7 & 50.1 & 62.8 \\

& \textcolor{tableblue}{Ours (T)}
& \bfseries 42.51 & 6.40 & 1.07
& \textcolor{tableblue}{Ours (T)}
& 68.8 & 60.6 & \cellcolor{tablefocus}\bfseries 50.5 & \bfseries 64.5 \\

\midrule

ARCTIC (s05, ego)
& InterField-SF & 29.65 & 9.81 & 0.82
& HACO & 42.2 & \bfseries 82.2 & 42.2 & 55.8 \\

& InterField-LSTM & 29.68 & 9.77 & \bfseries 0.76
& EgoPHI & 33.5 & 45.2 & 31.8 & 38.5 \\

& \textcolor{tableblue}{Ours (SF)}
& 20.02 & 10.58 & 1.09
& \textcolor{tableblue}{Ours (SF)}
& 59.2 & 67.3 & 50.9 & 63.0 \\

& \textcolor{tableblue}{Ours (T)}
& \bfseries 16.23 & \cellcolor{tablefocus}\bfseries 9.32 & 1.10
& \textcolor{tableblue}{Ours (T)}
& \bfseries 64.0 & 63.7 & \cellcolor{tablefocus}\bfseries 51.0 & \bfseries 63.8 \\

\midrule

DexYCB
& InterField-SF & 128.99 & 32.83 & 1.76
& HACO & 27.4 & \bfseries 85.5 & 34.3 & 41.5 \\

& InterField-LSTM & 127.43 & 31.05 & \bfseries 1.18
& EgoPHI & 4.4 & 4.2 & 2.7 & 4.3 \\

& \textcolor{tableblue}{Ours (SF)}
& 26.14 & 3.92 & 1.74
& \textcolor{tableblue}{Ours (SF)}
& 63.6 & 71.4 & 58.5 & 67.3 \\

& \textcolor{tableblue}{Ours (T)}
& \bfseries 23.29 & \cellcolor{tablefocus}\bfseries 3.42 & 1.46
& \textcolor{tableblue}{Ours (T)}
& \bfseries 65.7 & 69.7 & \cellcolor{tablefocus}\bfseries 59.5 & \bfseries 67.6 \\

\midrule

MOW
& InterField-SF & 108.13 & 33.15 & {--}
& HACO & 29.2 & \bfseries 71.5 & \cellcolor{tablefocus}\bfseries 34.5 & 41.5 \\

& InterField-LSTM & 110.08 & 35.00 & {--}
& ContactPrompt & 25.2 & 66.2 & 29.7 & 36.5 \\

& \textcolor{tableblue}{Ours (SF)}
& \bfseries 46.86 & \cellcolor{tablefocus}\bfseries 12.21 & {--}
& EgoPHI & 22.5 & 39.4 & 12.4 & 28.7 \\

& & & &
& \textcolor{tableblue}{Ours (T)}
& \bfseries 44.2 & 44.3 & \cellcolor{tablefocus}\bfseries 34.5 & \bfseries 44.3 \\

\bottomrule
\end{tabular}}
\end{table}

\Needspace{6\baselineskip}
For proximity, InterField-SF predicts distances per frame, whereas InterField-LSTM aggregates temporal features; both are trained on ARCTIC~\citep{fan2023arctic}. InterField covers 100\,mm and TactoHand is supervised up to 250\,mm, so we report full-range ADE and ADE-100, which clips both predictions and targets at 100\,mm.

TactoHand lowers full-range ADE across all four benchmarks: the temporal variant performs best on ARCTIC, DexYCB, and HOCap, and the single-frame variant on MOW. The latter also leads on HOCap ADE-100. ADE-100 separates accuracy within the shared distance range from the methods' different full ranges.

Turning to contact, the temporal model consistently improves Macro-F1 over HACO on DexYCB, HOCap, and ARCTIC, with the largest gain on HOCap, from 18.0 to 50.5. On MOW, TactoHand matches HACO in Macro-F1 (34.5) while improving Micro-F1 from 41.5 to 44.3. Figure~\ref{fig:qualitative-contact} illustrates representative missed and spurious contact regions across methods.

\Needspace{7\baselineskip}
\subsection{Force and Rollout Ablation}

We next evaluate the effect of motion-level physics supervision. We train single-frame and temporal variants with and without the rollout loss while holding all other training settings fixed. Table~\ref{tab:force-baselines} reports both vertex-force errors and a reference-assisted rigid-body rollout diagnostic. Appendix~\ref{app:force-rollout-evaluation} details the evaluation protocol and window-exclusion criteria.

\begin{table}[H]
    \centering
    \footnotesize
    \setlength{\tabcolsep}{3.5pt}
    \caption{
    Force and rollout ablation on DexYCB (DINOv3 ViT-L). SF/T: single-frame/temporal; MAE: mN; rollout ADE: cm; angles: degrees. Active targets exceed $0.05$\,N. The $0.4$\,s rollout uses reference-assisted intervals and excludes divergent windows (Appendix~\ref{app:force-rollout-evaluation}); bold indicates column minima.
    }
    \label{tab:force-baselines}
    \begin{tabular}{@{}lcccccc@{}}
        \toprule
        Model
        & Physics
        & Vertex MAE $\downarrow$
        & Active MAE $\downarrow$
        & Force angle $\downarrow$
        & Rollout ADE $\downarrow$
        & Rollout angle $\downarrow$ \\
        \midrule
        SF
        & Off
        & \textbf{12.6}
        & 124.9
        & 28.2
        & 5.3
        & 45.5 \\

        SF
        & On
        & 12.9
        & 126.7
        & 29.8
        & 3.8
        & \textbf{27.4} \\

        \addlinespace[2pt]

        T
        & Off
        & 13.7
        & 126.4
        & 31.0
        & 5.7
        & 45.2 \\

        T
        & On
        & 12.8
        & \textbf{122.7}
        & \textbf{27.5}
        & \cellcolor{tablefocus}\textbf{3.3}
        & 29.4 \\

        \bottomrule
    \end{tabular}
\end{table}

\Needspace{4\baselineskip}
Physics supervision primarily improves motion consistency: for the temporal model, rollout ADE falls from 5.7 to 3.3\,cm and active-force MAE from 126.4 to 122.7\,mN. The single-frame model also lowers rollout error, despite slightly worse active-force MAE and force-angle error.

\subsection{Downstream Hand--Object Reconstruction Refinement}
\label{sec:reconstruction}

We replace the original contact objective in EasyHOI~\citep{liu2025easyhoi} and ArtHOI~\citep{wang2026arthoi} with terms derived from fixed TactoHand contact and proximity predictions, keeping all other settings unchanged. This controlled setting uses metric object geometry; EasyHOI also uses ground-truth object orientation (Appendix~\ref{app:evaluation-details}).

On DexYCB, HOCap, and MOW, TactoHand improves PA-VPE in all six method--dataset pairs and gives the lowest hand--object Chamfer distance in each (Table~\ref{tab:reconstruction}). Full metrics and visuals appear in Appendix Table~\ref{tab:reconstruction-full} and Figures~\ref{fig:HOI-reconstruction}--\ref{fig:HOI-ablation}.

\begin{table}[H]
\centering
\footnotesize
\caption{
Controlled reconstruction refinement. PA-VPE: mm; $\mathrm{CD}_{\mathrm{ho}}$: bidirectional mean squared hand--object Chamfer in $10^{-3}\,\mathrm{m}^{2}$. Bold marks each method--dataset best.
}
\label{tab:reconstruction}
\setlength{\tabcolsep}{8pt}
\renewcommand{\arraystretch}{1.0}
\begin{tabular}{@{}llrrrrrr@{}}
\toprule
& &
\multicolumn{3}{c}{PA-VPE $\downarrow$}
&
\multicolumn{3}{c}{$\mathrm{CD}_{\mathrm{ho}}$ $\downarrow$} \\
\cmidrule(lr){3-5}
\cmidrule(l){6-8}
Dataset
& Method
& Base
& + HACO
& + Ours
& Base
& + HACO
& + Ours \\
\midrule

DexYCB
& EasyHOI
& 9.937
& 9.516
& \cellcolor{tablefocus}\textbf{9.077}
& 24.140
& 12.552
& \cellcolor{tablefocus}\textbf{2.004} \\

& ArtHOI
& 7.317
& 6.027
& \cellcolor{tablefocus}\textbf{4.626}
& 11.587
& 13.627
& \cellcolor{tablefocus}\textbf{1.431} \\

\midrule

HOCap
& EasyHOI
& 9.179
& 9.051
& \cellcolor{tablefocus}\textbf{7.592}
& 59.416
& 68.943
& \cellcolor{tablefocus}\textbf{11.461} \\

& ArtHOI
& 6.984
& 6.684
& \cellcolor{tablefocus}\textbf{6.073}
& 5.234
& 5.199
& \cellcolor{tablefocus}\textbf{1.104} \\

\midrule

MOW
& EasyHOI
& 14.832
& 14.276
& \cellcolor{tablefocus}\textbf{13.939}
& 69.080
& 72.638
& \cellcolor{tablefocus}\textbf{15.266} \\

& ArtHOI
& 13.572
& 13.261
& \cellcolor{tablefocus}\textbf{13.100}
& 73.993
& 73.147
& \cellcolor{tablefocus}\textbf{32.413} \\

\bottomrule
\end{tabular}
\end{table}

\section{Limitations}
\label{sec:limitations}
Geometry-derived contact is a proxy sensitive to source geometry and alignment. Motion-derived vertex forces are non-unique estimates, not measurements, and depend on inertia, friction, and environmental support assumptions. Motion-level evaluation is limited to short-horizon, reference-assisted rigid-body rollouts. Future work should incorporate measured forces and articulated rollout supervision while reducing reliance on auxiliary physical inputs.

\section{Conclusion}
\label{sec:conclusion}

TactoHand predicts contact, proximity, and 3D force as dense interaction fields on the hand surface. Geometry-derived labels across 24 sources, supplemented with regularized motion-derived force targets where available, support heterogeneous training. In our evaluations, TactoHand improves contact and proximity estimation, and its contact and proximity fields improve controlled hand--object reconstruction. Physics supervision reduces trajectory error in the reference-assisted rigid-body evaluation. Together, these results show how a shared hand-surface representation connects visual prediction, object motion, and hand--object reconstruction.

\section*{Reproducibility Statement}

The main text defines the hand-surface fields, label construction, model, and evaluation. Appendices~\ref{app:dataset-and-labels}--\ref{app:evaluation-details} document data sources and labels, additional results, model and training settings, and evaluation protocols. We will release the code and derived annotations to support reproduction of the labels and trained model.

\section*{AI Use Statement}

Generative AI tools assisted with retrieval and discovery of related work, drafting sections of the paper, and polishing its writing. The authors take responsibility for the final content, including its claims, citations, figures, and results.

\section*{Ethics Statement}

This work uses only previously released datasets and collects no new human-subject data. The sources were assembled by their original authors under their own consent and licensing arrangements, and we neither re-identify participants nor add identifying information. We record each dataset's source and preprocessing steps so that derived annotations can be traced to the original data. Users must comply with the terms of each source dataset.

The labels we derive are estimates, not measurements. Vertex forces are obtained from smoothed object motion through inverse dynamics and a constrained allocation, and the construction admits many local force fields for a single aggregate wrench; the dataset records which frames satisfy the force-allocation constraints and never substitutes a zero-force target for an infeasible one. Monocular force estimation is also dual-use, in that an accurate estimate of applied force could inform physical manipulation or contact-based control. We therefore report force labels with their validity flags and residuals, and we caution against treating them, or any force prediction trained on them, as ground truth in safety-critical settings.

\bibliography{iclr2027_conference}
\bibliographystyle{iclr2027_conference}

\clearpage
\appendix
\raggedbottom
\makeatletter\setlength{\@fptop}{0pt}\makeatother
\section*{Appendix Roadmap}
The appendix presents dataset composition and label construction (Appendix~\ref{app:dataset-and-labels}), additional results (Appendix~\ref{app:additional-results}), model and training details (Appendix~\ref{app:implementation-details}), and evaluation protocols and coverage (Appendix~\ref{app:evaluation-details}).
\section{Dataset and Label Construction}
\label{app:dataset-and-labels}

\subsection{Dataset Inventory}
\label{app:dataset-inventory}

Table~\ref{tab:dataset-inventory} lists the 24 curated sources and their label types. We derive contact and proximity on the canonical MANO surface from source geometry. Sources with motion-derived force labels are grouped by whether the object motion used to construct them is rigid or articulated.

Counts describe our converted collection, not the full dataset releases or the number of training samples. Frames are counted across all captured camera views. Except for ObMan and MOW ($\ast$), sources contain ordered sequences; for those two datasets, ``sequences'' are groups of independent images.

For GigaHands ($\dagger$), we select 15 objects because the remaining objects have less reliable tracking. The table counts all camera views of sequences containing at least one selected object, including frames in those sequences that are excluded from training. HOCap ($\ddagger$) contributes to the collection total but is excluded from training. PROX contributes 178 distinct source frames captured at 5\,Hz; the table reports this original frame count. HO3Dv2/v3 refer to HO-3D release versions, and OakInk-v1/v2 to OakInk and OakInk2.

\begingroup
\small
\renewcommand{\arraystretch}{1.08}
\begin{longtable}{@{}l@{\hspace{2em}}r@{\hspace{2em}}r@{}}
    \caption{Data sources and labels in the curated collection.}
    \label{tab:dataset-inventory}\\

    \toprule
    Dataset & Sequences & Frames \\
    \midrule
    \endfirsthead

    \multicolumn{3}{@{}l}{\emph{Table~\ref{tab:dataset-inventory} continued}} \\
    \toprule
    Dataset & Sequences & Frames \\
    \midrule
    \endhead

    \multicolumn{3}{@{}l}{\emph{Force labels: articulated objects}} \\
    ARCTIC~\citep{fan2023arctic} & 300 & 1\,903\,869 \\
    HOI4D~\citep{liu2022hoi4d} & 2\,973 & 891\,900 \\

    \addlinespace[2pt]
    \multicolumn{3}{@{}l}{\emph{Force labels: rigid objects}} \\
    GigaHands$^\dagger$~\citep{fu2025gigahands} & 1\,424 & 9\,449\,724 \\
    Show3D~\citep{rim2026show3d} & 1\,682 & 6\,931\,884 \\
    TACO~\citep{liu2024taco} & 2\,120 & 4\,287\,490 \\
    HOGraspNet~\citep{cho2024hograspnet} & 13\,575 & 4\,270\,708 \\
    OakInk-v2~\citep{zhan2024oakink2} & 611 & 3\,898\,190 \\
    HOT3D~\citep{banerjee2025hot3d} & 198 & 701\,519 \\
    HOCap$^\ddagger$~\citep{wang2025hocap} & 64 & 653\,175 \\
    DexYCB~\citep{chao2021dexycb} & 1\,000 & 581\,968 \\
    H2O~\citep{kwon2021h2o} & 184 & 453\,855 \\
    OakInk-v1~\citep{yang2022oakink} & 792 & 235\,604 \\
    HO3Dv3~\citep{hampali2020honnotate} & 17 & 103\,722 \\
    HO3Dv2~\citep{hampali2020honnotate} & 17 & 77\,953 \\
    H2O3D~\citep{hampali2022keypoint} & 15 & 67\,525 \\

    \addlinespace[2pt]
    \multicolumn{3}{@{}l}{\emph{Contact and proximity labels only}} \\
    InterHand2.6M~\citep{moon2020interhand} & 255\,378 & 12\,469\,860 \\
    ReInterHand~\citep{moon2023reinterhand} & 250 & 886\,470 \\
    DECAF~\citep{shimada2023decaf} & 16 & 778\,817 \\
    Hi4D~\citep{yin2023hi4d} & 1\,600 & 191\,120 \\
    ObMan$^\ast$~\citep{hasson2019learning} & 23 & 118\,133 \\
    RICH~\citep{huang2022bstro} & 111 & 62\,969 \\
    HIC~\citep{tzionas2016capturing} & 11 & 1\,824 \\
    MOW$^\ast$~\citep{cao2021reconstructing} & 512 & 512 \\
    PROX~\citep{hassan2019prox} & 1 & 178 \\

    \midrule
    \textbf{Total} & \textbf{282\,874} & \textbf{49\,018\,969} \\
    \bottomrule
\end{longtable}
\endgroup

Sources for which force labels can be constructed account for approximately $70\%$ of curated frames; this does not mean that every frame contributes force supervision. GigaHands contributes contact and proximity supervision, while its motion-derived force targets are omitted because of noise in object poses. HOCap is reserved for evaluation. The rollout loss uses eligible rigid-object episodes only; articulated episodes provide vertex-force supervision.

\subsection{Geometric Contact-Label Comparison}
\label{app:contact-label-comparison}

\begin{figure}[H]
\centering
\includegraphics[width=0.94\linewidth,trim=0bp 110bp 10bp 78bp,clip]{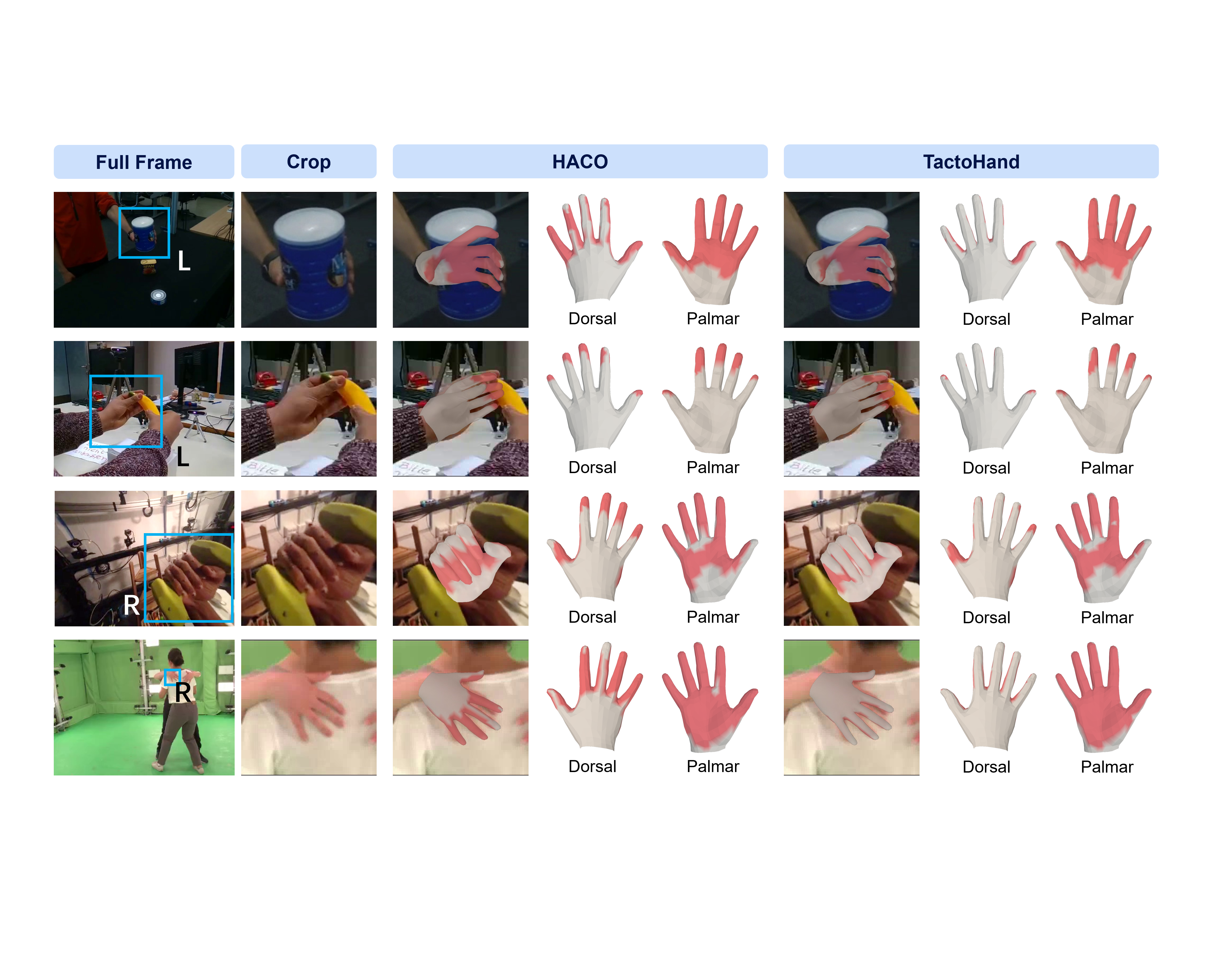}
\caption{Comparison of contact labels generated by HACO's distance-only rule and our geometric rule. Normal filtering removes back-facing false positives, while inside-mesh inclusion recovers penetrating vertices missed by distance thresholding.}
\label{fig:Annotation-Comparison}
\end{figure}

Following HACO~\citep{jung2025dense}, we use distance thresholds of $10$\,mm for hand--object sources, $35$\,mm for MOW, and $5$\,mm for InterHand2.6M, HIC, and ReInterHand; all remaining sources use $10$\,mm. Equation~\ref{eq:contact-decision} further applies normal filtering and inside-mesh inclusion, as illustrated in Figure~\ref{fig:Annotation-Comparison}.

\paragraph{Multiray inside--outside test.}
\label{app:inside-outside-test}

To determine whether a hand vertex lies inside a closed object mesh, we transform it into object coordinates and cast several oblique rays against the original, unrepaired triangles. For each ray, an odd number of intersections indicates that the vertex is inside; a majority vote across rays sets the sign of its nearest-surface distance. A small offset in the ray test avoids ambiguous intersections at edges and vertices. For open or unreliable surfaces, we retain the unsigned distance and skip inside-mesh inclusion.

\subsection{Motion-Derived Force Labels}
\label{app:force-label-details}

\subsubsection{Trajectory Processing}
\label{app:trajectory-processing}

\paragraph{Temporal sampling and smoothing.}

We interpolate observed object trajectories onto a fixed $30$\,Hz timeline for both rigid- and articulated-object force labeling. Translation is smoothed with a symmetric, truncated Gaussian kernel. For rotation, we project each input matrix onto $\mathrm{SO}(3)$ by SVD, then compute a weighted intrinsic mean. We initialize the rotation estimate at the center rotation and update it iteratively as follows:
\begin{equation}
\boldsymbol\delta^{(k)}_t
=
\sum_{j=-r_R}^{r_R} g^R_j
\operatorname{Log}\!\left(
    (\bar{\mathbf R}^{(k)}_t)^\top \mathbf R_{t+j}
\right),
\qquad
\bar{\mathbf R}^{(k+1)}_t
=
\bar{\mathbf R}^{(k)}_t
\operatorname{Exp}(\boldsymbol\delta^{(k)}_t),
\label{eq:intrinsic-rotation-smoothing}
\end{equation}
where $g^R_j$ are normalized Gaussian weights. We use eight iterations and bandwidths of $\sigma_p=0.075\,\mathrm{s}$ for translation and $\sigma_R=0.150\,\mathrm{s}$ for rotation. At solver rate $f$, either kernel has radius $r(\sigma,f)=\max\{\operatorname{round}(0.30f),\lceil 3\sigma f\rceil\}$ frames. We clamp indices outside the trajectory to the nearest endpoint when smoothing, but supervise a transition only when the full smoothing windows around both states fall within the observed trajectory.

\subsubsection{Rigid-Object Force Allocation and Validation}
\label{app:force-allocation-details}

\paragraph{Trajectory-derived target wrench.}

For a rigid object, let $(\bar{\mathbf x}_t,\bar{\mathbf R}_t)$ denote its smoothed center-of-mass pose and let $\Delta t=1/f$. The displacement over transition $t\rightarrow t+1$ gives the outgoing linear velocity:
\begin{equation}
\mathbf v_{t+1}
=
\frac{\bar{\mathbf x}_{t+1}-\bar{\mathbf x}_t}{\Delta t}.
\end{equation}
We obtain the outgoing world-frame angular velocity from $\bar{\mathbf R}_{t+1}\bar{\mathbf R}_t^\top$ using the Cayley map. With $\boldsymbol\phi_t=\operatorname{Log}(\bar{\mathbf R}_{t+1}\bar{\mathbf R}_t^\top)$ and $\theta_t=\lVert\boldsymbol\phi_t\rVert_2$,
\begin{equation}
\boldsymbol\omega^{w}_{t+1}
=
\frac{2\tan(\theta_t/2)}{\Delta t\,\theta_t}
\boldsymbol\phi_t,
\label{eq:cayley-angular-velocity}
\end{equation}
with the continuous limit $\boldsymbol\phi_t/\Delta t$ as $\theta_t\rightarrow0$. The incoming velocity is the preceding transition's outgoing velocity; for the first transition, we copy its outgoing velocity. Dividing the velocity change by $\Delta t$ yields the acceleration used in target wrench $\mathbf w_t^\star$. We express angular velocity and acceleration in the body frame when computing the torque in Equation~\ref{eq:inverse-dynamics-wrench}.

We obtain gravity direction from dataset metadata where available, UprightNet RGB estimates~\citep{xian2019uprightnet}, or a manually inferred scene or world-down axis. Camera-frame estimates are transformed into the solver world frame. The inverse-dynamics target uses a $1\,\mathrm{kg}$ reference mass and the corresponding scaled inertia, without rollout optimization.

\paragraph{Candidate vertices and grasp map.}

We consider force candidates for a hand--object pair only when at least one hand vertex is labeled as contact under the $10$\,mm rule in Section~\ref{sec:label-generation}. Within that pair, vertices with $s_{t,i}<20$\,mm are candidates, weighted by Equation~\ref{eq:force-candidate-priority}.

Let $\mathbf p_i$ and $\mathbf f_i$ denote the world position and hand-on-object force of an active vertex. The grasp map
\begin{equation}
\mathbf G_i\mathbf f_i
=
\begin{bmatrix}
\mathbf f_i\\
(\mathbf p_i-\mathbf x)\times\mathbf f_i
\end{bmatrix}
\label{eq:force-grasp-map}
\end{equation}
maps its force to a wrench about the object's center of mass $\mathbf x$. Candidate normals $\mathbf n_{t,i}$ are normalized, and each force is constrained to a six-facet unilateral Coulomb cone with $\mu=1.2$.

\paragraph{Environment support.}

In Equation~\ref{eq:force-allocation}, $e_t\geq0$ denotes upward support at an estimated tabletop contact point, and $\mathbf G_t^{\mathrm{env}}$ maps it to a wrench about the object center of mass. For tabletop sequences, we estimate this point from mesh vertices within $2$\,mm of the object's lowest point. We infer whether the object rests on the table or is held above it by comparing its height with nearby hand-free frames. For each hand-contact interval, median heights from up to 15 such frames on either side provide reference heights before and after contact. Before identifying these intervals, we bridge gaps of at most five solver frames in the hand-contact signal. A rise of more than $7$\,mm for three frames marks pickup; a sustained return to the post-contact reference height marks placement. If the object is already held at the start of an interval, we use the post-contact reference when available; without reference frames, we treat it as off-table.

These support states are estimates rather than observed labels. We accept a frame without nearby active hand force only if table support alone can explain its motion-derived target wrench. When hand and environment contact coexist, the QP jointly allocates hand forces and nonnegative support at the estimated point.

\paragraph{QP implementation.}

We solve Equation~\ref{eq:force-allocation} as a sparse sequence QP. The force-magnitude, force-temporal, and environment-temporal coefficients default to one; the environment-magnitude coefficient and priority denominator $\epsilon$ are $10^{-4}$. Torque equations are scaled by $0.05\ \mathrm{N\,m}$. Temporal differences use the union of adjacent active-vertex sets, treating absent vertices as zero. OSQP~\citep{stellato2020osqp} uses absolute and relative tolerances of $10^{-6}$. Partitioning, cached factorizations, and warm starts leave the objective unchanged.

\paragraph{Validity and replay.}

We reconstruct the fitted wrench from the solved vertex and environment forces and record its force and torque residuals. Independently, we replay that wrench from the stored smoothed state in Newton Physics~\citep{newtonphysics2025} and compare the one-step pose with the next smoothed state.

\Needspace{6\baselineskip}
\subsubsection{Articulated-Object Force Labels}
\label{app:articulated-force-labels}

For an articulated object with $J$ joints, we retain its kinematic tree on the same $30$\,Hz timeline. We distribute the $1$\,kg reference mass among links in proportion to mesh volume and compute each link's center of mass and inertia at its assigned mass. Multibody inverse dynamics on the smoothed trajectory gives the required generalized effort. Subtracting the generalized gravity contribution yields the contact target
\begin{equation}
\boldsymbol\eta_t^{\star}
=
\boldsymbol\eta_t^{\mathrm{ID}}
-
\boldsymbol\eta_t^{g}
\in\mathbb R^{6+J},
\label{eq:articulated-target}
\end{equation}
whose first six components are the floating-base force and torque, expressed in the solver world frame with torque about the system center of mass; the remaining components are joint efforts. The target can be computed from smoothed base and joint trajectories or, equivalently, by projecting per-link Newton--Euler wrenches onto these generalized coordinates.

We assign each candidate hand vertex to its nearest object part. A force $\mathbf f_{t,i}$ applied at $\mathbf p_{t,i}$ contributes $\mathbf f_{t,i}$ to the base force and $(\mathbf p_{t,i}-\mathbf c_t)\times\mathbf f_{t,i}$ to torque about the system center of mass $\mathbf c_t$. For a joint $j$ ancestral to the contacted part, its joint-effort contribution is
\begin{equation}
\gamma_{t,i,j}
=
\begin{cases}
\mathbf a_{t,j}^{\top}\!\left(
    (\mathbf p_{t,i}-\mathbf o_{t,j})\times\mathbf f_{t,i}
\right), & \text{revolute},\\
\mathbf a_{t,j}^{\top}\mathbf f_{t,i}, & \text{prismatic},
\end{cases}
\label{eq:articulated-contact-map}
\end{equation}
where $\mathbf a_{t,j}$ and $\mathbf o_{t,j}$ are the joint axis and pivot in world coordinates. Joints outside the part's ancestor chain receive zero contribution. This part-aware map replaces the rigid grasp map in a sequence-level allocation with the same signed-distance priority, unilateral friction constraints, and temporal force penalty.

Permitted environment support contributes to base balance. Joint effort unexplained by contact forces remains an internal, non-hand contribution instead of being assigned to hand vertices. Hand-free frames have zero hand force. If a hand-contact allocation fails, we record that the force estimate is invalid or that environment support was used as a fallback, rather than treating the failed estimate as a valid hand force. We check fitted generalized effort against its target and use valid articulated vertex forces for field-level supervision, without the rigid rollout loss.

\FloatBarrier
\section{Additional Results and Visualizations}
\label{app:additional-results}
\setlength{\intextsep}{6pt}

\begin{figure}[H]
\centering
\includegraphics[width=0.95\linewidth,trim=0bp 110bp 10bp 78bp,clip]{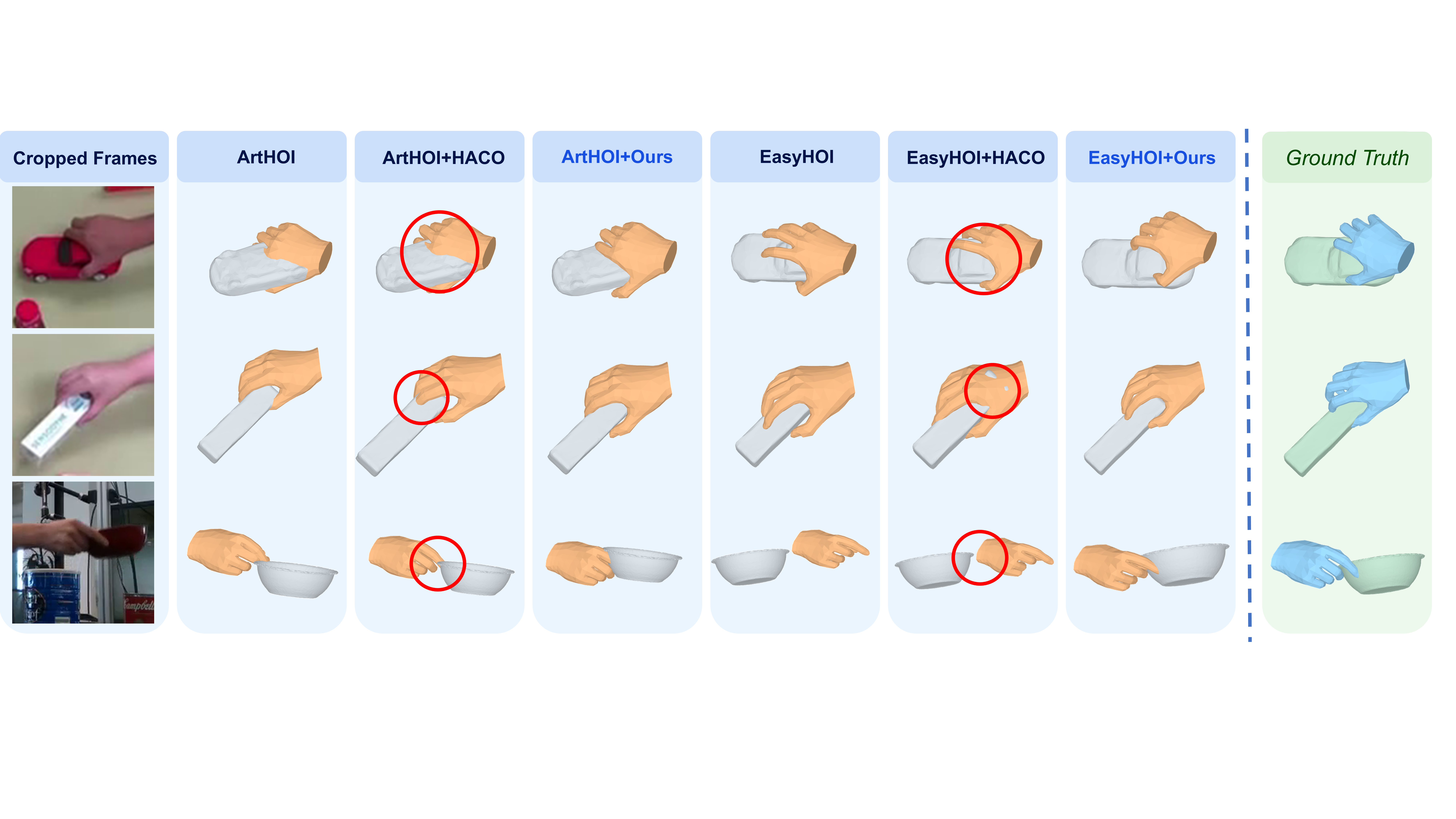}
\caption{HOCap~\citep{wang2025hocap} reconstructions. Red circles highlight hand--object contact artifacts in the HACO variants.}
\label{fig:HOI-reconstruction}
\end{figure}

\begin{figure}[H]
\centering
\includegraphics[width=0.95\linewidth,trim=80bp 110bp 138bp 78bp,clip]{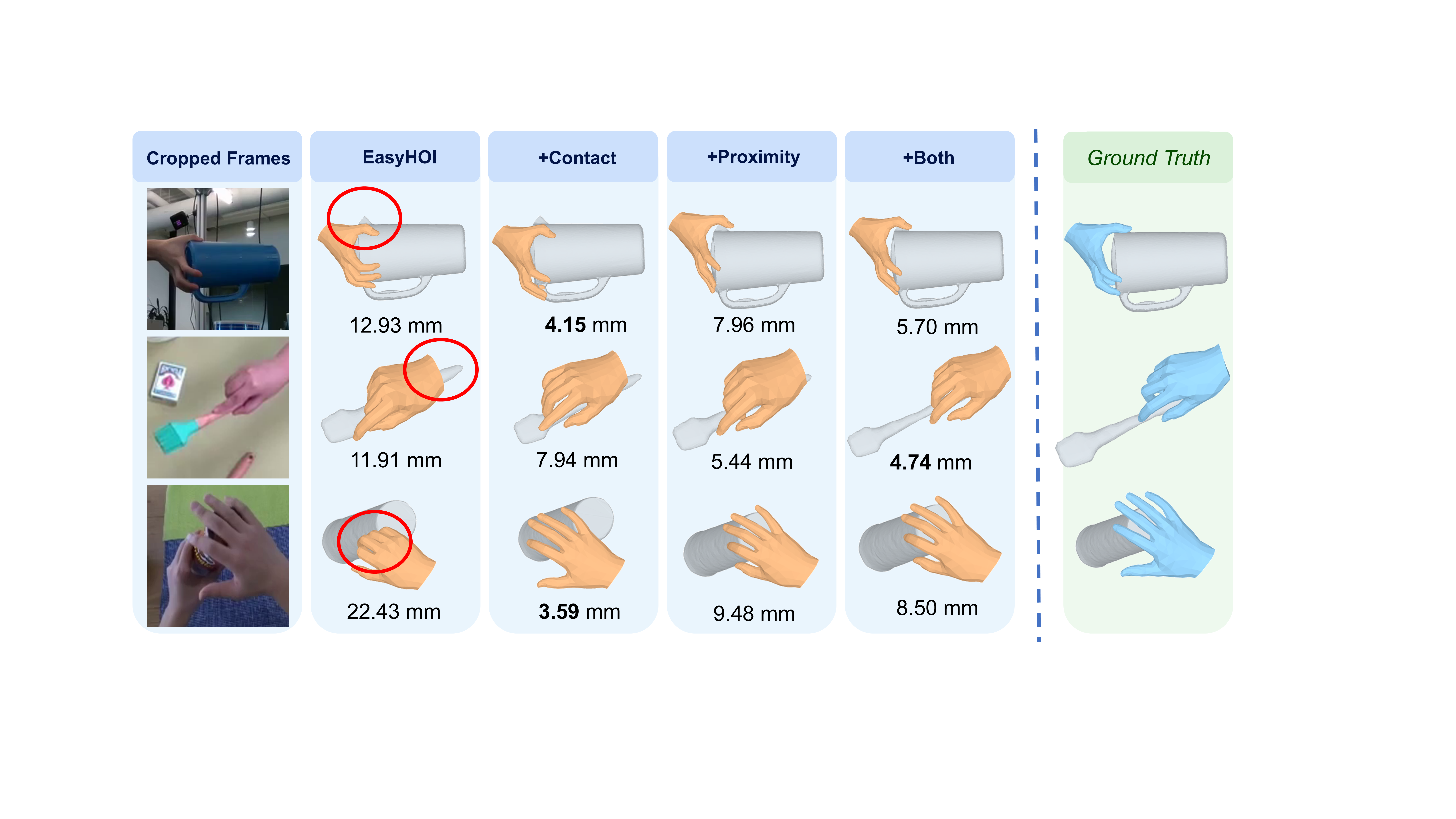}
\caption{EasyHOI~\citep{liu2025easyhoi} reconstruction ablation. + Contact: contact-normal term; + Proximity: distance term; + Both: both terms. Values are per-example PA-VPE (mm).}
\label{fig:HOI-ablation}
\end{figure}

Table~\ref{tab:reconstruction-full} extends the main reconstruction comparison with joint errors and hand--object F-scores. Across all six method--dataset pairs, TactoHand also improves these metrics over the base and HACO variants. Figure~\ref{fig:HOI-reconstruction} illustrates the resulting hand--object alignment on HOCap, while Figure~\ref{fig:HOI-ablation} separates the effects of the contact and proximity terms on individual examples.

\begin{table}[!htbp]
\centering
\footnotesize
\caption{Full reconstruction results. PA-VPE/JPE: mm; $\mathrm{CD}_{\mathrm{ho}}$: $10^{-3}\,\mathrm{m}^{2}$; F-5/F-10: 5/10\,mm; ACC: $\mathrm{mm}/\mathrm{s}^{2}$. Bold marks the best per method--dataset block; shading highlights PA-VPE and $\mathrm{CD}_{\mathrm{ho}}$.}
\label{tab:reconstruction-full}
\setlength{\tabcolsep}{3pt}
\renewcommand{\arraystretch}{0.72}
\begin{tabular}{@{}l S[table-format=2.3] S[table-format=2.3] S[table-format=2.3] S[table-format=1.4] S[table-format=1.4] S[table-format=4.1]@{}}
\toprule
Methods & \multicolumn{1}{c}{PA-VPE $\downarrow$} & \multicolumn{1}{c}{PA-JPE $\downarrow$} & \multicolumn{1}{c}{$\mathrm{CD}_{\mathrm{ho}}$ $\downarrow$} & \multicolumn{1}{c}{$\mathrm{F\text{-}5}_{\mathrm{ho}}$ $\uparrow$} & \multicolumn{1}{c}{$\mathrm{F\text{-}10}_{\mathrm{ho}}$ $\uparrow$} & \multicolumn{1}{c}{ACC $\downarrow$} \\
\midrule
\multicolumn{7}{c}{DexYCB} \\
\midrule
EasyHOI & 9.937 & 10.483 & 24.140 & 0.3609 & 0.5359 & \multicolumn{1}{c}{--} \\
EasyHOI + HACO & 9.516 & 10.032 & 12.552 & 0.3609 & 0.5392 & \multicolumn{1}{c}{--} \\
EasyHOI + Ours & \cellcolor{tablefocus}\bfseries 9.077 & \bfseries 9.559 & \cellcolor{tablefocus}\bfseries 2.004 & \bfseries 0.3817 & \bfseries 0.5606 & \multicolumn{1}{c}{--} \\
ArtHOI & 7.317 & 7.520 & 11.587 & 0.3930 & 0.5240 & 4311.6 \\
ArtHOI + HACO & 6.027 & 6.190 & 13.627 & 0.3970 & 0.5300 & 3978.9 \\
ArtHOI + Ours & \cellcolor{tablefocus}\bfseries 4.626 & \bfseries 4.747 & \cellcolor{tablefocus}\bfseries 1.431 & \bfseries 0.4540 & \bfseries 0.6200 & \bfseries 3892.7 \\
\midrule
\multicolumn{7}{c}{HOCap} \\
\midrule
EasyHOI & 9.179 & 9.790 & 59.416 & 0.2306 & 0.3991 & \multicolumn{1}{c}{--} \\
EasyHOI + HACO & 9.051 & 9.623 & 68.943 & 0.2216 & 0.3824 & \multicolumn{1}{c}{--} \\
EasyHOI + Ours & \cellcolor{tablefocus}\bfseries 7.592 & \bfseries 8.075 & \cellcolor{tablefocus}\bfseries 11.461 & \bfseries 0.2456 & \bfseries 0.4319 & \multicolumn{1}{c}{--} \\
ArtHOI & 6.984 & 7.243 & 5.234 & 0.3215 & 0.5085 & 5176.9 \\
ArtHOI + HACO & 6.684 & 6.892 & 5.199 & 0.3249 & 0.5118 & 4666.3 \\
ArtHOI + Ours & \cellcolor{tablefocus}\bfseries 6.073 & \bfseries 6.264 & \cellcolor{tablefocus}\bfseries 1.104 & \bfseries 0.3672 & \bfseries 0.5869 & \bfseries 4599.7 \\
\midrule
\multicolumn{7}{c}{MOW} \\
\midrule
EasyHOI & 14.832 & 15.860 & 69.080 & 0.2262 & 0.3728 & \multicolumn{1}{c}{--} \\
EasyHOI + HACO & 14.276 & 15.215 & 72.638 & 0.2195 & 0.3617 & \multicolumn{1}{c}{--} \\
EasyHOI + Ours & \cellcolor{tablefocus}\bfseries 13.939 & \bfseries 14.715 & \cellcolor{tablefocus}\bfseries 15.266 & \bfseries 0.2370 & \bfseries 0.3888 & \multicolumn{1}{c}{--} \\
ArtHOI & 13.572 & 14.054 & 73.993 & 0.2330 & 0.3640 & \multicolumn{1}{c}{--} \\
ArtHOI + HACO & 13.261 & 13.727 & 73.147 & 0.2340 & 0.3670 & \multicolumn{1}{c}{--} \\
ArtHOI + Ours & \cellcolor{tablefocus}\bfseries 13.100 & \bfseries 13.545 & \cellcolor{tablefocus}\bfseries 32.413 & \bfseries 0.2610 & \bfseries 0.4060 & \multicolumn{1}{c}{--} \\
\bottomrule
\end{tabular}
\end{table}

\FloatBarrier
\section{Model and Training Details}
\label{app:implementation-details}

\subsection{Network Architecture}
\label{app:network-architecture}

A trainable DINOv3 ViT-L encodes each $256{\times}256$ RGB frame independently. We average its final patch tokens, excluding class and register tokens, to obtain one 1024-dimensional feature vector per frame ($D=1024$ in Figure~\ref{fig:pipeline}). A six-layer temporal transformer with eight attention heads processes these features with learned positional embeddings. A shared two-layer, eight-head query decoder processes three learned queries, one each for force, contact, and proximity. At each image timestamp, each decoded query is fused with the corresponding frame feature to predict its field for both hands. Transformer feed-forward layers have width $4D$, and dropout is zero.

The four variants combine two temporal modes (SF and T) with rollout supervision (Off and On). T processes the full clip through the temporal transformer and shared query decoder. SF applies the same modules to each frame as a length-one sequence, preserving the architecture and output shapes while removing cross-frame attention. All variants share the force, contact, and proximity head designs; On enables the rollout loss, whereas Off sets its weight to zero.

\subsection{Force Coordinates, Temporal Sampling, and Supervision}
\label{app:field-clock-validity}

\paragraph{Vertex-local force coordinates.}

Figure~\ref{fig:force-allocation} (lower left) shows the local basis used by the force head. Its normal is the outward normal of the posed MANO mesh; one tangent points approximately toward the wrist, and the other completes an orthonormal basis. We expand the predicted force components along these axes in camera coordinates before force supervision and wrench aggregation. Contact and proximity retain canonical vertex indexing.

\paragraph{Vertex-force loss.}

Before computing the camera-frame force loss, we cap predicted and target magnitudes at $20$\,N by rescaling each vector without changing its direction. We then separate targets by capped magnitude: active targets above $0.05$\,N receive a vertex-weighted Huber penalty with a $0.05$\,N transition, while the remaining targets receive mean squared error weighted by $0.1$.

\Needspace{15\baselineskip}
\paragraph{Differentiable rigid-body rollout.}

The model predicts fields at 10\,Hz. Each 100\,ms visual interval spans three 30\,Hz force intervals: the first uses predicted vertex forces, and the next two use detached reference forces. Each force interval contains four 120\,Hz integration steps of size $h=1/120$\,s, following Equation~\ref{eq:rollout-state-update}. The object state propagates continuously, without resets to ground-truth poses.

For one step, let $(\mathbf x,\mathbf v,\mathbf q,\boldsymbol\omega^w)$ denote the center-of-mass position, linear velocity, orientation quaternion, and world-frame angular velocity, and let $(\mathbf F,\boldsymbol\tau)$ be the net world-frame wrench about the center of mass. Define $\mathbf R=\mathbf R(\mathbf q)$, $\boldsymbol\omega^b=\mathbf R^\top\boldsymbol\omega^w$, $\boldsymbol\tau^b=\mathbf R^\top\boldsymbol\tau$, and $\bar{\boldsymbol\omega}^b=(\boldsymbol\omega^b+\boldsymbol\omega^{b,+})/2$. With body-frame inertia $\mathbf I$, the update is
\begin{equation}
\begin{aligned}
\mathbf v^+
    &= \mathbf v + \tfrac{h}{m}\mathbf F,\\
\mathbf x^+
    &= \mathbf x + h\mathbf v^+,\\[2pt]
\boldsymbol\omega^{b,+}
    &= \boldsymbol\omega^b
       + h\mathbf I^{-1}
       \bigl(
           \boldsymbol\tau^b
           -\bar{\boldsymbol\omega}^b
            \times\mathbf I\bar{\boldsymbol\omega}^b
       \bigr),\\
\boldsymbol\omega^{w,+}
    &= \mathbf R\boldsymbol\omega^{b,+},\\
\mathbf q^+
    &= \operatorname{normalize}\!\left(
       \mathcal Q(h\boldsymbol\omega^{w,+})
       \otimes\mathbf q
       \right).
\end{aligned}
\label{eq:rollout-state-update}
\end{equation}
We solve the body-frame angular-velocity update with six unrolled Newton iterations. The pre-update rotation $\mathbf R$ then maps the result to the world frame for the orientation update; $\mathcal Q$ maps an axis-angle vector to a unit quaternion. All state updates are differentiable, allowing the rollout loss to backpropagate through wrench aggregation to predicted vertex forces. Each visual interval contains one force interval driven by predictions; the other two use fixed reference forces while preserving gradients through the evolving object state.

\paragraph{Supervision eligibility.}

The available annotations determine which losses a sequence contributes to. Geometry-only sequences contribute contact and proximity losses. Rigid and articulated sequences with force labels also contribute force loss at image timestamps where those labels are valid. A field-level loss is evaluated only when the corresponding hand and label are available, regardless of whether that hand is occluded in the image. Invalid force labels do not contribute to the force loss.

Rigid rollout further requires a trajectory whose hand force can be attributed to the target object. We exclude a trajectory if, at any force-valid time step, the hand applies nonzero force to another object while the target receives none. Repeated boundary frames used to complete a visual clip do not contribute to force or rollout losses; both use only time intervals within the recorded sequence.

\subsection{Training Configuration}

The four variants share the final-stage settings in Table~\ref{tab:training-config}. Dataset sampling weights are proportional to the number of available 32-frame windows, multiplied by the listed dataset-specific factors.

\begin{table}[!htbp]
\centering
\small
\caption{Training settings and dataset sampling multipliers. Unlisted training sources use a multiplier of $1$.}
\label{tab:training-config}
\setlength{\tabcolsep}{5pt}
\begin{tabular}{@{}p{0.22\linewidth}p{0.68\linewidth}@{}}
\toprule
Setting & Value \\
\midrule
Hardware and batch
    & One node with eight A800 GPUs; 7 samples per GPU; two accumulation steps (112 samples per update) \\
Optimizer
    & AdamW; learning rate $2\times10^{-5}$ for backbone and heads; weight decay $10^{-4}$; cosine decay with two-epoch warmup \\
Window sampling
    & 32 frames, stride 32; each epoch samples 20\% of the windows allocated to each dataset \\
Rollout-loss weight
    & $10^{-4}$ with physics; $0$ without physics \\
Dataset sampling multipliers
    & $4$: MOW; $2$: DexYCB; $0.5$: HOI4D, OakInk-v2; $0.1$: DECAF, ReInterHand; $0.05$: HOGraspNet, ObMan, GigaHands; $0.02$: InterHand2.6M \\
\bottomrule
\end{tabular}
\end{table}

Each 150-epoch final-stage run initializes model weights from an earlier checkpoint but resets the optimizer, scheduler, and epoch state. In Equation~\ref{eq:capability-objective}, $\lambda_c=1$, $\lambda_d=2$, and $\lambda_p=10$, with $\mathcal L_{\mathrm{contact}}=\mathcal L_{\mathrm{conditional}}+0.25\mathcal L_{\mathrm{engagement}}$. No separate direct wrench loss is used. For variants with physics supervision, $\lambda_r$ increases from zero in the second epoch to $10^{-4}$ in the sixth epoch; it remains zero for variants without physics supervision. In Equation~\ref{eq:rollout-pose-loss}, $\mathcal H=\{1,\ldots,12\}$ spans four 10\,Hz intervals, $\lambda_x=[2(0.01\,\mathrm m)^2]^{-1}$, and $\lambda_R=[2(0.1\,\mathrm{rad})^2]^{-1}$. Rollout losses are also averaged over eligible windows. At test time, contact is predicted when $\hat c>0.5$.

We report the checkpoints with the best validation performance. Our validation sets are the nominal DexYCB test split and the ARCTIC P2 validation split, and the reported results are measured on these same sets. HOCap is excluded from training. Its results characterize cross-dataset transfer.

\FloatBarrier
\section{Evaluation Protocol and Coverage}
\label{app:evaluation-details}

\subsection{Contact and Proximity Comparisons}
Table~\ref{tab:prox_contact} re-scores released checkpoints on our geometry-derived annotations and nominal dataset splits. Input pipelines can differ. SF/T denote single-frame/temporal fusion; a dash denotes an unavailable result or undefined acceleration for static MOW. MOW uses SF for proximity and T for contact. For MOW, we form the temporal input by repeating each image across the 32-frame clip; no additional temporal observations are available.

\paragraph{Metric definitions.}
We decode predicted proximity $\widehat r$ as $\widehat d=-\log(\widehat r)/\tau$ in metres and report mean absolute distance error (ADE) in mm over valid vertices. ADE-100 clips both prediction and target at 100\,mm before computing the error; all valid vertices remain in the average. This matches InterField's 100\,mm output range, whereas TactoHand's target extends to 250\,mm. For distance error $e_{t,i}=\widehat d_{t,i}-d_{t,i}$, Accel. averages $|e_{t+1,i}-2e_{t,i}+e_{t-1,i}|$ multiplied by the squared frame-sampling rate, in m/s$^2$. It measures temporal variation in distance error. We exclude any three-frame group containing a repeated frame inserted to pad a clip.

For Macro-F1, we compute F1 over the vertices of each hand in each frame, then average over frame--hand pairs with contact in either the ground truth or the prediction. Failing to predict any ground-truth contact on a hand, or predicting contact on a hand with none, gives a score of zero. Pairs with no contact in either are excluded from this average. Micro-F1 pools vertex-level true positives (TP), false positives (FP), and false negatives (FN) over all valid frame--hand pairs and computes $2TP/(2TP+FP+FN)$. Correctly predicted non-contact vertices do not increase either score.

\paragraph{Input protocols.}
ARCTIC uses its official P2 validation split (subject s05, egocentric view, all labeled frames); DexYCB, HOCap, and MOW use their nominal test splits. On ARCTIC, InterField uses its released P2 egocentric checkpoints with the official preprocessing of raw fisheye images and crops around the interaction bounding box. Its reported ADE against its original reference labels is 8.83\,mm for SF and 8.40\,mm for LSTM. On DexYCB, HOCap, and MOW, InterField instead uses P1 allocentric checkpoints on hand crops, making these out-of-domain evaluations. EgoPHI is trained only on ARCTIC and uses its released weights and deployed contact threshold of 0.3. Following its training protocol, it uses native full frames except for MOW, where it uses hand crops. TactoHand uses full-frame images on DexYCB, HOCap, and ARCTIC, and hand crops on MOW, where the hand occupies a small fraction of the frame.

\subsection{Downstream Reconstruction Protocol}
We evaluate TactoHand's contact and proximity fields in EasyHOI~\citep{liu2025easyhoi} and ArtHOI~\citep{wang2026arthoi} on DexYCB~\citep{chao2021dexycb}, HOCap~\citep{wang2025hocap}, and MOW~\citep{cao2021reconstructing}. DexYCB and MOW are included in training; HOCap is excluded. MOW contains independent images, while the other two provide video. Within each reconstruction method, variants share the model, initialization, images, and optimization budget. Table~\ref{tab:reconstruction-full} compares base, HACO~\citep{jung2025dense}, and TactoHand variants; Figure~\ref{fig:HOI-ablation} isolates the EasyHOI contact and proximity terms.

\paragraph{Shared interaction terms.}
We hold the predicted contact and proximity values at the 778 MANO vertices fixed throughout reconstruction refinement. Proximity is decoded as $\widehat d=-\log(\widehat r)/\tau$ with $\tau=20\,\mathrm{m}^{-1}$; values saturated at $0.25\,\mathrm{m}$ specify a lower bound on distance. The proximity term applies a Huber penalty ($\delta=1$) to signed distance residuals normalized by $5\,\mathrm{mm}$. In the contact-normal term, predicted contact probability weights the factor $\max(0,\mathbf{n}_h^\top\mathbf{n}_o+0.5)^2$, where $\mathbf n_h$ and $\mathbf n_o$ denote the hand and object surface normals. The proximity and contact-normal terms both have coefficient $1$. They replace the original contact objective, while reconstruction, penetration, and regularization terms remain.

\begin{table}[!htbp]
\centering
\small
\caption{Reconstruction stages and optimization settings shared across the compared variants.}
\label{tab:reconstruction-stages}
\setlength{\tabcolsep}{3pt}
\begin{tabular}{@{}p{0.21\linewidth}p{0.11\linewidth}p{0.35\linewidth}p{0.25\linewidth}@{}}
\toprule
Stage & Updates & Optimized variables & Learning rate \\
\midrule
ArtHOI alignment & 250 & Hand translation & $10^{-2}\!\to\!5\times10^{-4}$ \\
ArtHOI hand refinement & 399 & Finger pose, hand translation & $10^{-3}\!\to\!10^{-4}$ \\
ArtHOI penetration correction & 300 & Hand translation, global orientation & $10^{-3}$ \\
ArtHOI smoothing & 300 & Hand translation & $5\times10^{-4}$ \\
\midrule
EasyHOI global alignment & $20\times100$ & Hand-global parameters, object translation & $5\times10^{-2}$ \\
EasyHOI hand refinement & 1,000 & Pose residual, object translation; MANO shape & $10^{-2}$; $10^{-4}$ for shape \\
\bottomrule
\end{tabular}
\end{table}

\paragraph{EasyHOI refinement.}
We fix metric CAD geometry and ground-truth object orientation, optimizing translation from its image-based initialization. Table~\ref{tab:reconstruction-stages} lists 3,000 updates, excluding image-based initialization and auxiliary PCA fitting, with constant learning rates within each stage. Hand refinement starts from the aligned hand parameters, without backpropagating through the preceding alignment stage; the hand's root rotation remains fixed during refinement. During global alignment, interaction losses are enabled under the same conditions as in the original method. The proximity term ramps up over the first 50 updates in which those losses are evaluated, while the contact-normal term is off. During refinement, proximity is fully enabled and the contact-normal term ramps up over 10 such updates.

\paragraph{ArtHOI refinement.}
We fix metric object geometry, skip scale optimization, and use Adam with $(\beta_1,\beta_2)=(0.9,0.999)$. Table~\ref{tab:reconstruction-stages} lists the 1,249 refinement updates performed after initialization. For the first two stages, learning rates are taken from step 151 onward of the original 800-step schedule. An L1 penalty with coefficient $0.4$ discourages deviation from the initial pose. The proximity term ramps up over 50 alignment iterations, and the contact-normal term over 10 hand-refinement iterations. Both remain active during penetration correction; only proximity remains during smoothing.

PA-VPE and PA-JPE are Procrustes-aligned mean vertex and joint errors, respectively, in mm. $\mathrm{CD}_{\mathrm{ho}}$ sums bidirectional mean squared nearest-neighbor distances in $10^{-3}\,\mathrm{m}^{2}$; $\mathrm{F\text{-}5}_{\mathrm{ho}}$ and $\mathrm{F\text{-}10}_{\mathrm{ho}}$ are hand--object F-scores at 5 and 10\,mm. ACC is acceleration error in $\mathrm{mm}/\mathrm{s}^{2}$, available only for ArtHOI on DexYCB and HOCap; dashes in Table~\ref{tab:reconstruction-full} mean unavailable. These are reconstruction outcomes, not direct contact or proximity scores.

\subsection{Force and Rollout Evaluation}
\label{app:force-rollout-evaluation}
Table~\ref{tab:force-baselines} evaluates four single-frame/temporal $\times$ physics-off/on variants on the DexYCB test split. They share data, backbone, and optimization settings (Appendix~\ref{app:network-architecture}); physics-on adds rigid rollout loss. Force metrics compare predicted and motion-derived reference vectors in the camera frame at image timestamps with valid labels, excluding repeated frames used for padding. Vertex MAE averages component-wise absolute error over all supervised vertices, including zero-force targets. Active MAE and force angle use only targets above $0.05$\,N; angle is the mean predicted-to-reference vector angle. MAEs are in mN and angles in degrees.

Rollout uses a 0.4\,s horizon: four 10\,Hz visual intervals or twelve 30\,Hz force intervals. Within each visual interval, predicted forces drive the first force interval and detached reference forces drive the next two; four 120\,Hz updates integrate each force interval. Sliding windows start from observed object states and include specified gravity and estimated environment-support forces. We score only intervals within the recorded sequence, excluding those added to pad a clip. Position ADE averages object-center displacement error over the horizon, and rollout angle averages rotation geodesic error. A window is excluded if any state is non-finite or its linear or angular speed exceeds $100$\,m/s or $1000$\,rad/s, respectively. Reported errors therefore describe the non-divergent windows of this reference-assisted rollout.

\end{document}